\documentclass[final]{cvpr}

\usepackage{times}
\usepackage{epsfig}
\usepackage{graphicx}
\usepackage{amsmath}
\usepackage{amssymb}
\usepackage{enumerate}

\usepackage{url}

\usepackage{microtype}
\usepackage{graphicx}
\usepackage{subfigure}
\usepackage{booktabs} 
\usepackage{bm}
\usepackage{multirow}
\usepackage{balance}

\usepackage{algorithm}  
\usepackage{algorithmic}

\usepackage{footnote}

\newcommand{\model}{\textsc{Face}\textsc{Sec}}
\DeclareMathOperator*{\argmax}{arg\,max}

\newcommand{\nop}[1]{}

\usepackage[pagebackref=true,breaklinks=true,colorlinks,bookmarks=false]{hyperref}

\def\cvprPaperID{5477} 
\def\confYear{CVPR 2021}

\begin{document}

\title{\textsc{Face}\textsc{Sec}: A Fine-grained Robustness Evaluation Framework \\for Face Recognition Systems}


\author{Liang Tong$^{1,2}$\thanks{Work done during an internship at NEC Laboratories America.}, 
Zhengzhang Chen$^2$\thanks{Corresponding author.}, 
Jingchao Ni$^2$, 
Wei Cheng$^2$,\\
Dongjin Song$^3$, 
Haifeng Chen$^2$, 
Yevgeniy Vorobeychik$^1$ \\
$^1$Washington University in St. Louis, {\tt\small\{liangtong, yvorobeychik\}@wustl.edu} \\
$^2$NEC Laboratories America, {\tt\small\{zchen,jni,weicheng,haifeng\}@nec-labs.com} \\
$^3$University of Connecticut, {\tt\small dongjin.song@uconn.edu} \\
}


\maketitle

\begin{abstract}

We present \model, a framework for fine-grained robustness evaluation of face recognition systems.
\model\ evaluation is performed along four dimensions of adversarial modeling: the nature of perturbation (\textit{e.g.}, pixel-level or face accessories), the attacker's system knowledge (about training data and learning architecture), goals (dodging or impersonation), and capability (tailored to individual inputs or across sets of these).
We use \model\ to study five face recognition systems in both closed-set and open-set settings, and to evaluate the state-of-the-art approach for defending against physically realizable attacks on these. 
We find that accurate knowledge of neural architecture is significantly more important than knowledge of the training data in black-box attacks. 
Moreover, we observe that open-set face recognition systems are more vulnerable than closed-set systems under different types of attacks. 
The efficacy of attacks for other threat model variations, however, appears highly dependent on both the nature of perturbation and the neural network architecture. For example, attacks that involve adversarial face masks are usually more potent, even against adversarially trained models, and the ArcFace architecture tends to be more robust than the others.

\end{abstract}
\section{Introduction}
\label{sec:introduction}

Face recognition has received much attention~\cite{jain2011handbook,parkhi2015deep,schroff2015facenet,wen2016discriminative,liu2017sphereface,wang2018cosface} in recent years. Empowered by deep convolutional neural networks (CNNs), it has become widely used in various areas, including security-sensitive applications, such as airport check-in, online financial transactions, and mobile device login. 
The success of such \emph{deep face recognition} is particularly striking, with $>$99\% prediction accuracy on benchmark datasets~\cite{parkhi2015deep, liu2016large,liu2017sphereface, deng2019arcface}.

Despite its widespread success in computer vision applications, recent studies have found that 
deep face recognition models are vulnerable to \emph{adversarial examples} in both \emph{digital space}~\cite{madry2018towards, dong2019efficient, yang2020delving} and \emph{physical space}~\cite{sharif2016accessorize}.
The former directly modifies an input face image by adding imperceptible perturbations to mislead face recognition (henceforth, \emph{digital attacks}).
The latter is characterized by adding adversarial perturbations that can be realized on \emph{physical objects} (\textit{e.g.}, wearing an adversarial eyeglass frame~\cite{sharif2016accessorize}), which are subsequently captured by a camera and then fed into a face recognition model to fool prediction (henceforth, \emph{physically realizable attacks}).
As such, the aforementioned domains, especially critical domains such as security and finance, are subjected to risks of opening the backdoor for the attackers. 
For example, in face recognition supported financial/banking services, an illegal user may bypass biometric verification and steal money from victims' accounts. 
Therefore, there exists a vital need for methods that can comprehensively and systematically evaluate the robustness of face recognition systems in adversarial settings, which in turn can shed light on the design of robust models for downstream tasks.


The main challenges of comprehensive evaluation of the robustness of face recognition lie in dealing with the diversity of face recognition systems and adversarial environments. First, different face recognition systems consist of various key components (\textit{e.g.}, training data and neural architecture); such diversity results in different performance and robustness.
To enable comprehensive and systematic evaluations, it is crucial 
to assess the robustness of every individual or a combination of face recognition components in adversarial settings. Second, adversarial example attacks can vary by the nature of perturbations (\textit{e.g.}, pixel-level or physical space), an attacker's goal, knowledge, and capability. For a given face recognition system, its robustness against a specific type of attack may not generalize to other kinds~\cite{wu2019defending}.

In spite of recent advances in adversarial attacks~\cite{sharif2016accessorize, dong2019efficient, yang2020delving} that demonstrate the vulnerability of face recognition systems, most existing methods fail to address the aforementioned challenges 
due to the following reasons. 
First, current efforts appeal to either \emph{white-box attacks} or \emph{black-box attacks} to obtain a lower bound or upper bound of robustness.
These bounds indicate the vulnerability of face recognition systems in adversarial settings but lack the understanding of how each component of face recognition contributes to such vulnerability.
Second, while most existing approaches focus on a specific type of attack (\textit{e.g.}, digital attacks that incur imperceptible noise~\cite{dong2019efficient, yang2020delving}), they fail to explore the different levels of robustness in response to various attacks (\textit{e.g.}, physically realizable attacks).



To bridge this gap, we propose \model, a fine-grained robustness evaluation framework for face recognition systems.
\model\ incorporates four dimensions in evaluation: the nature of adversarial perturbations (pixel-level or face accessories), the attacker's accurate knowledge about the target face recognition system (training data and neural architecture), goals (dodging or impersonation), and capability (individual or universal attacks).
Specifically, we implement both digital and physically realizable attacks in \model. We leverage the PGD attack~\cite{madry2018towards}, the state-of-the-art digital attack paradigm, and the eyeglass frame attack~\cite{sharif2016accessorize} as the representative of physically realizable attacks. Additionally, we propose two novel physically realizable attacks: one involves pixel-level adversarial stickers on human faces, and the other adds color grids on face masks. 
Moreover, to facilitate universal attacks that produce \emph{image-agnostic} perturbations, we propose a systematic approach that works on top of the attack paradigms described above. 

In summary, this paper makes the following contributions:
\begin{enumerate}[(1)]
\vspace{-5pt}
\item We propose \model, the first robustness evaluation framework that enables researchers to (i) identify the vulnerability of each face recognition component to adversarial examples, and (ii) assess different levels of robustness under various adversarial circumstances.
\vspace{-5pt}
\item We propose two novel physically realizable attacks: the pixel-level sticker attack and the grid-level face mask attack. These allow us to explore adversarial robustness against different types of physically realizable perturbations. Particularly, the latter responds to the pressing needs for security analysis of face recognition systems, as face masks have become common face accessories during the COVID-19 pandemic.
\vspace{-5pt}
\item We propose a general approach to produce universal adversarial examples for a batch of face images. 
Compared to previous works, our paradigm has a significant speedup and is more efficient in evaluation. 
\vspace{-5pt}
\item We perform a comprehensive evaluation on five publicly available face recognition systems in various settings to demonstrate the efficacy of \model. 
\end{enumerate} 

\section{Background and Related Work}
\label{sec:backgrond}


\subsection{Face Recognition Systems}

\begin{figure}[t]
\vspace{-5pt}
\centering
\includegraphics[width=0.5\textwidth]{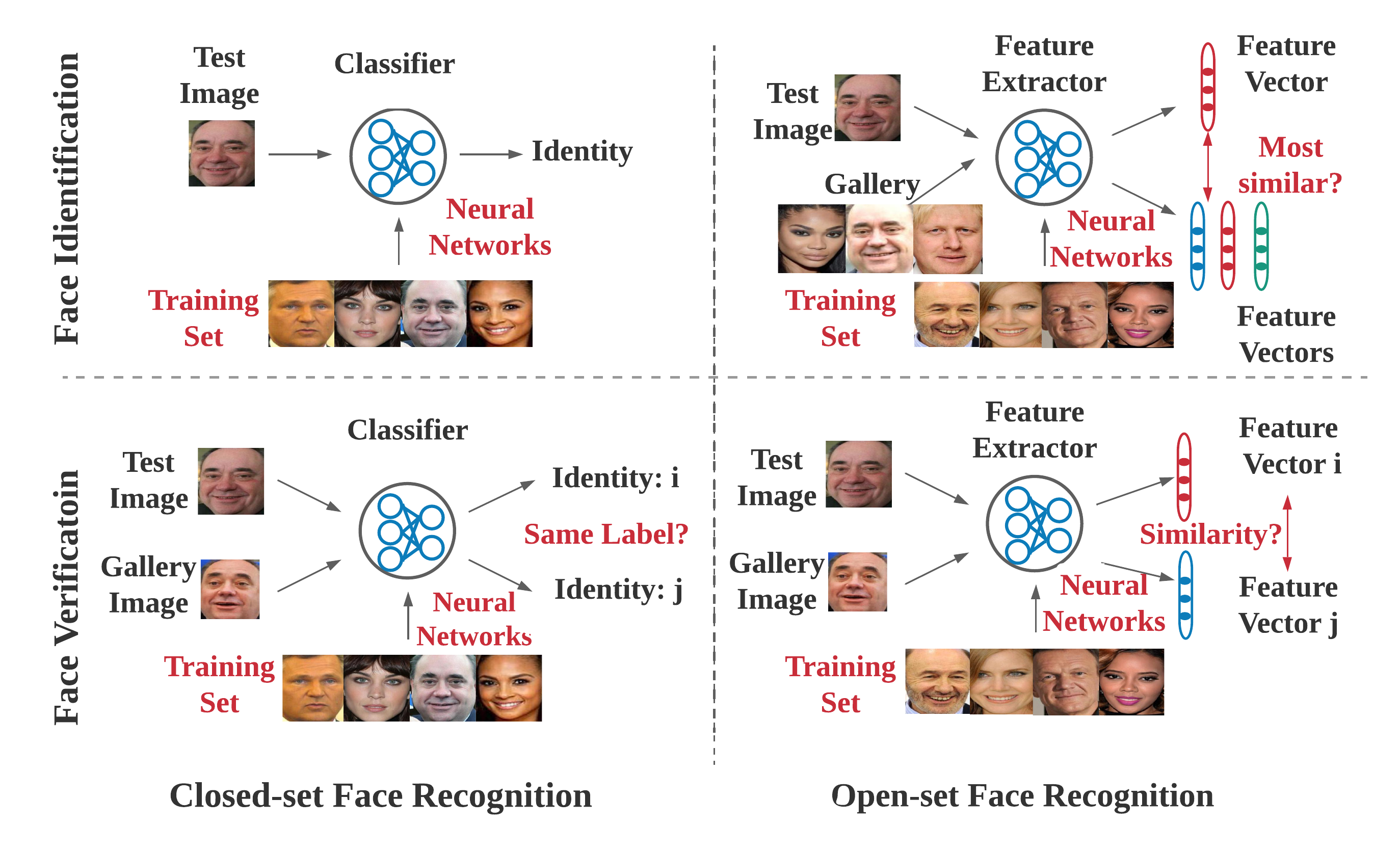} 
\caption{Closed-set and open-set face recognition systems.}
\label{fig:face_recognition}
\vspace{-5pt}
\end{figure}


Generally, deep face recognition systems aim to solve the following two tasks:
1) \emph{Face identification}, which returns the predicted identity of a test face image;
2) \emph{Face verification}, which indicates whether a test face image (also called probe face image) and the face image stored in the gallery belong to the same identity.
Based on whether all testing identities are predefined in the training set, face recognition systems can be further categorized into \emph{closed-set systems} and \emph{open-set systems}~\cite{liu2017sphereface}, as illustrated in Fig.~\ref{fig:face_recognition}. 

In closed-set face recognition tasks, all the testing samples' identities are enrolled in the training set. Specifically, a face identification task is equivalent to a \emph{multi-class classification} problem by using the standard softmax loss function in the training phase~\cite{taigman2014deepface,sun2014deepa,sun2014deepb}. And a face verification task is a natural extension of face identification by first performing the classification twice (one for the test image and the other for the gallery) and then comparing the predicted identities to see if they are identical.

In contrast, there are usually no overlaps between identities in the training and testing set for open-set tasks. In this setting, a face verification task is essentially a \emph{metric learning} problem, which aims to maximize \emph{intra-class distance} and minimize \emph{inter-class distance} under a chosen metric space by two steps~\cite{schroff2015facenet, parkhi2015deep,wen2016discriminative,liu2016large,liu2017sphereface,deng2019arcface}. First, we train a feature extractor that maps a face image into a discriminative feature space by using a carefully designed loss function; Then, we measure the distance between feature vectors of the test and gallery face images to see if it is above a verification threshold. As an extension of face verification, the face identification task requires additional steps to compare the distances between the feature vectors of the test image and each gallery image, and then choose the gallery's identity corresponding to the shortest distance.

This paper focuses on face identification for closed-set systems, as face verification is just an extension of identification in this setting.
Likewise, we focus on face verification for open-set systems.

\subsection{Digital and Physical Adversarial Attacks}


Recent studies have shown that deep neural networks are vulnerable to adversarial attacks. These attacks produce \emph{imperceptible} perturbations on images in the digital space to mislead classification~\cite{szegedy14intriguing, goodfellow15explaining, carlini2017towards} (henceforth, \emph{digital attacks}). 
While a number of attacks on face recognition fall into this category (\textit{e.g.}, by adding small $\ell_p$ bounded noise over the entire input~\cite{dong2019efficient} or perceptible but semantically meaningful perturbation on a restricted area of the input~\cite{qiu2019semanticadv}), of particular interest in face recognition, are attacks in the physical world (henceforth, \emph{physical attacks}). 


Generally, physical attacks have three characteristics~\cite{wu2019defending}.
First, the attackers directly modify the actual entity rather than digital features. 
Second, the attacks can mislead state-of-the-art face recognition systems.
Third, the attacks have low suspiciousness (\textit{i.e.}, by adding objects similar to common ``noise'' on a small part of human faces).
For example, an attacker can fool a face recognition system by wearing an adversarial eyeglass frame~\cite{sharif2016accessorize}, a standard face accessory in the real world.  

\begin{figure}
\centering
\begin{tabular}{ccc}
     \includegraphics[width=0.1\textwidth]{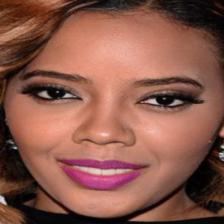} &
     \includegraphics[width=0.1\textwidth]{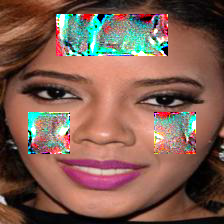} &
     \includegraphics[width=0.1\textwidth]{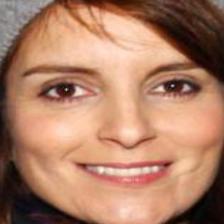} \\
\end{tabular}
\caption{Sticker attack: an example of physically realizable attacks on face recognition systems. 
Left: original input image.  Middle: adversarial sticker on the face.  Right: predicted identity.
In practice, the adversarial stickers can be printed and put on human faces.}
\label{fig:realizable_attack}
\vspace{-8pt}
\end{figure}

In this paper, we focus on both digital attacks and \emph{the digital representation of physical attacks} (henceforth, \emph{physically realizable attacks}).
Specifically, physically realizable attacks are digital attacks that can produce adversarial perturbations with low suspiciousness, and these perturbations can be realized in the physical world by using techniques such as 3-D printing (\textit{e.g.}, Fig. \ref{fig:realizable_attack} illustrates one example of such attacks on face recognition systems).
Compared to physical attacks, physically realizable attacks can evaluate robustness of face recognition systems more efficiently:
on the one hand, realizable attacks allow us to iteratively modify digital images directly so the evaluation can significantly speedup compared to modifying real-world objects and then photographing them;
on the other hand, robustness to physically realizable attacks provides the lower bound of robustness to physical attacks, as the former has fewer constraints and larger solution space.

Formally, both digital and physically realizable attacks can be performed by solving the following general form of an optimization problem (\textit{e.g.}, for closed-set identification task): 
\vspace{-5pt}
\begin{equation}
\argmax_{\bm{\delta}} \ell (S(\bm{x}+M\bm{\delta}), y) \ \ \ \ \ \ \ s.t.\  \bm{\delta} \in \Delta,
\label{eq:adv_example}
\vspace{-1pt}
\end{equation}
where $S$ is the target face recognition model, $\ell$ is the adversary's utility function (\textit{e.g.}, the loss function used to train $S$), $\bm{x}$ is the original input face image, $y$ is the associated identity, $\bm{\delta}$ is the adversarial perturbation, and $\Delta$ is the feasible space of the perturbation.
Here, $M$ denotes the mask matrix that constrains the area of perturbation; it has the same dimension as $\bm{\delta}$ and contains $1$s where perturbation is allowed, and $0$s where there is no perturbation.  

\subsection{Adversarial Defense for Face Recognition}
\label{subsec:defense}
While there have been numerous defense approaches to make face recognition robust to adversarial attacks, many of them focus on digital attacks and have been proved to be broken under adaptive attacks~\cite{carlini2017towards, tramer2020adaptive}.
Here, we describe one representative defense approach, \emph{adversarial training}~\cite{madry2018towards}, that is scalable, not defeated by adaptive attacks, and has been leveraged to defend against physically realizable attacks on face recognition systems.

The main idea of adversarial training is to minimize prediction loss of the training data, where an attacker tries to maximize the loss.
In practice, this can be done by iteratively using the following two steps: 1) Use an attack method to produce adversarial examples of the training data;
2) Use any optimizer to minimize the loss of predictions on these adversarial examples.
Wu \textit{et al.}~\cite{wu2019defending} propose to use DOA---adversarial training with the rectangular occlusion attacks---to defend against physically realizable attacks on closed-set face recognition systems.
Specifically, the rectangular occlusion attack included in DOA first heuristically locates a rectangular area among a collection of possible regions in an input face image, then fixes the position and adds adversarial occlusion inside the rectangle.
It has been shown that DOA can significantly improve the robustness against the eyeglass frame attack~\cite{sharif2016accessorize} for closed-set VGG-based face recognition system~\cite{parkhi2015deep} by 80\%.
However, as we will show in Section~\ref{sec:experiments}, DOA would fail to defend against other types of attacks, such as the face mask attack proposed in Section~\ref{subsec:perturbation}. 
\section{Methodology}
\label{sec:methodology}

\begin{figure}[t]
\vspace{-5pt}
\centering
 \includegraphics[width=0.45\textwidth]{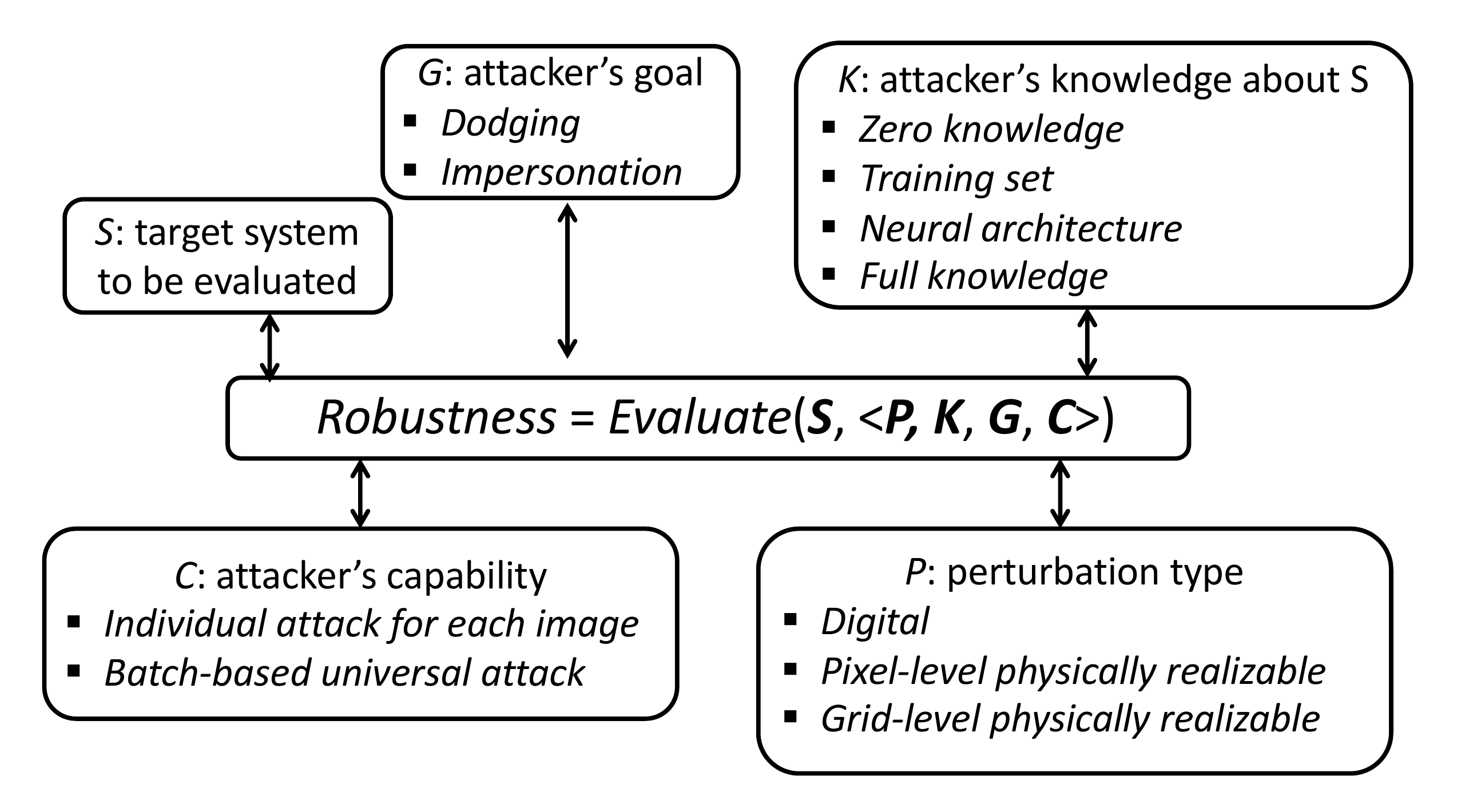} 
\caption{An overview of \model.}
\label{fig:methodoloy}
\vspace{-5pt}
\end{figure}

In this section, we introduce \model\ for fine-grained robustness evaluation of face recognition systems. Our goal is twofold: 1) identify vulnerability/robustness of each essential component that comprises a face recognition system, and 2) assess robustness in a variety of adversarial settings. 
Fig.~\ref{fig:methodoloy} illustrates an overview of \model. 
Let $S = f(h;D)$ be a face recognition system with a neural architecture $h$ that is trained on a training set $D$ by an algorithm $f$ (\textit{e.g.}, stochastic gradient descent), \model\ evaluates the robustness of $S$ via a quadruplet:
\vspace{-2pt}
\begin{equation}
    \mathsf{Robustness} = \mathsf{Evaluate}(S, <\mathit{P, K, G, C}>),
\vspace{-2pt}
\end{equation}
where $<P, K, G, C>$ represents an attacker who tries to produce adversarial examples to fool $S$. 
$P$ is the perturbation type, such as perturbations produced by pixel-level digital attacks and physically realizable attacks.
$K$ denotes the attacker's knowledge on the target system $S$, 
{\em i.e.}, the information about which sub-components of $S$ are leaked to the attacker.
$G$ is 
the goal of the attacker, such as the circumvention of detection and the misrecognition as a target identity. 
$C$ represents the attacker's capability.
For example, an attacker can either 
individually perturb each input face image, or produce universal perturbations for images batch-wise.
Next, we will describe each element of \model\ in details.

\subsection{Perturbation Type (P)} 
\label{subsec:perturbation}

\begin{figure}[t]
\centering
\includegraphics[width=0.45\textwidth]{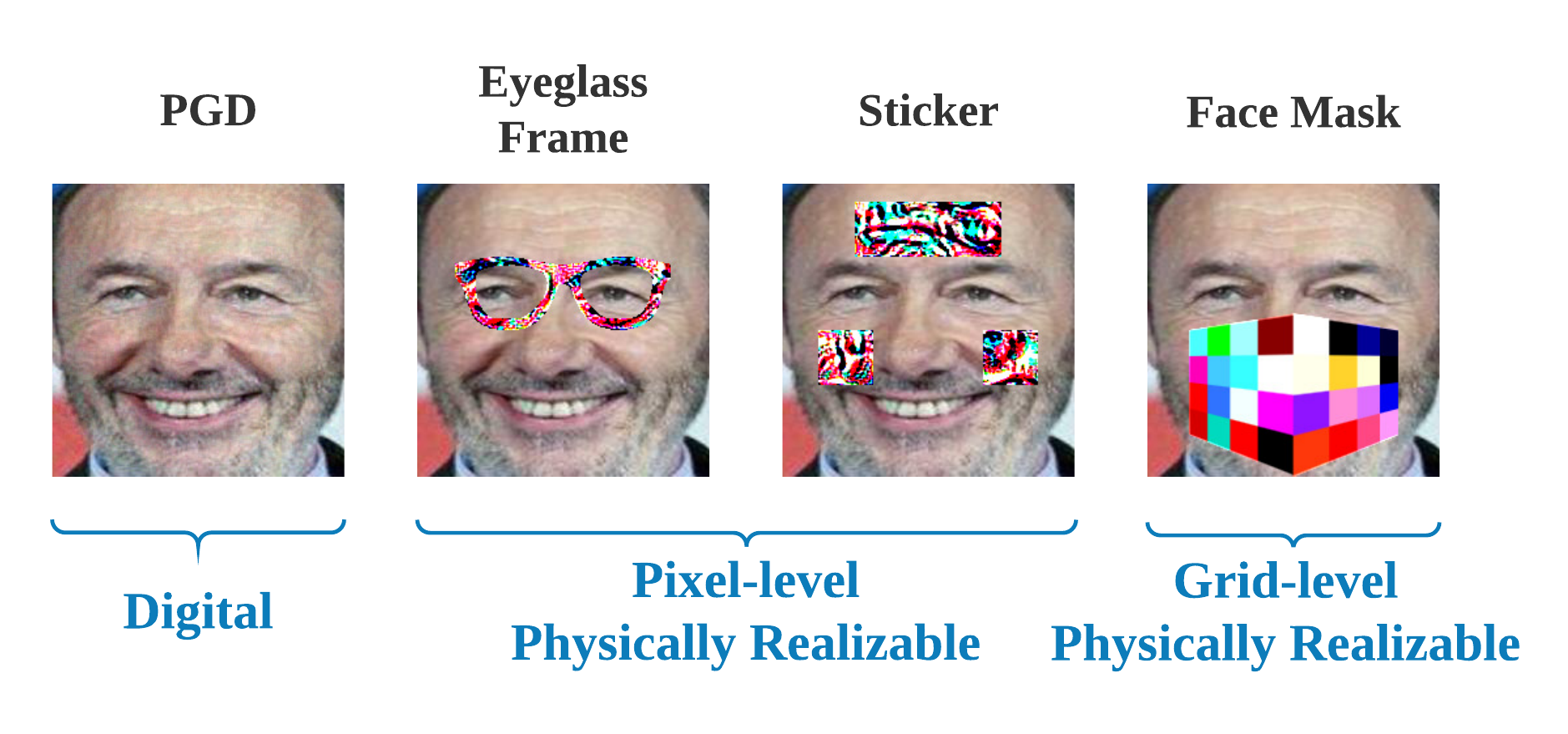} 
\caption{Perturbation types in \model.}
\label{fig:perturbation_type}
\vspace{-5pt}
\end{figure}

In \model, we consider three categories of attacks with different perturbation types: \emph{digital attack}, \emph{pixel-level physically realizable attack}, and \emph{grid-level physically realizable attack}, as shown in Fig.~\ref{fig:perturbation_type}.

\vspace{0.1cm}
\noindent{\bf Digital Attack}.
Digital attack produces small perturbations on the entire input face image.
We use the $\ell_\infty$-norm version of the PGD attack~\cite{madry2018towards} as the representative of this category\footnote{We also tried other digital attacks (\textit{e.g.}, CW~\cite{carlini2017towards} and JSMA~\cite{papernot2016limitations}), but these were either less effective than PGD or unable to be extended to universal attacks (see Section~\ref{subsec:capability}).}.

\vspace{0.1cm}
\noindent{\bf Pixel-level Physically Realizable Attack}.
This category of attack features pixel-level perturbations that can be realized in the physical world (\textit{e.g.}, by printing them on glossy photo papers).
In this case, the attacker adds large pixel-level perturbations on a small area of the input image (\textit{e.g.}, face accessories).
In \model, we use two attacks of this category: \emph{eyeglass frame attack}~\cite{sharif2016accessorize} and \emph{sticker attack}.
The former allows large perturbations within an eyeglass frame, and it can successfully mislead VGG-based face recognition systems~\cite{parkhi2015deep}.
We propose the latter to produce pixel-level perturbations that are added on less important face areas than the eyeglass frame, \textit{i.e.}, the two cheeks and forehead of human faces, as illustrated in Fig.~\ref{fig:realizable_attack} and \ref{fig:perturbation_type}.
Typically, the stickers are rectangular occlusions, which cover a total of about $20\%$ area of an input face image.   

\begin{figure}[t]
\vspace{-5pt}
\centering
 \includegraphics[width=0.38\textwidth]{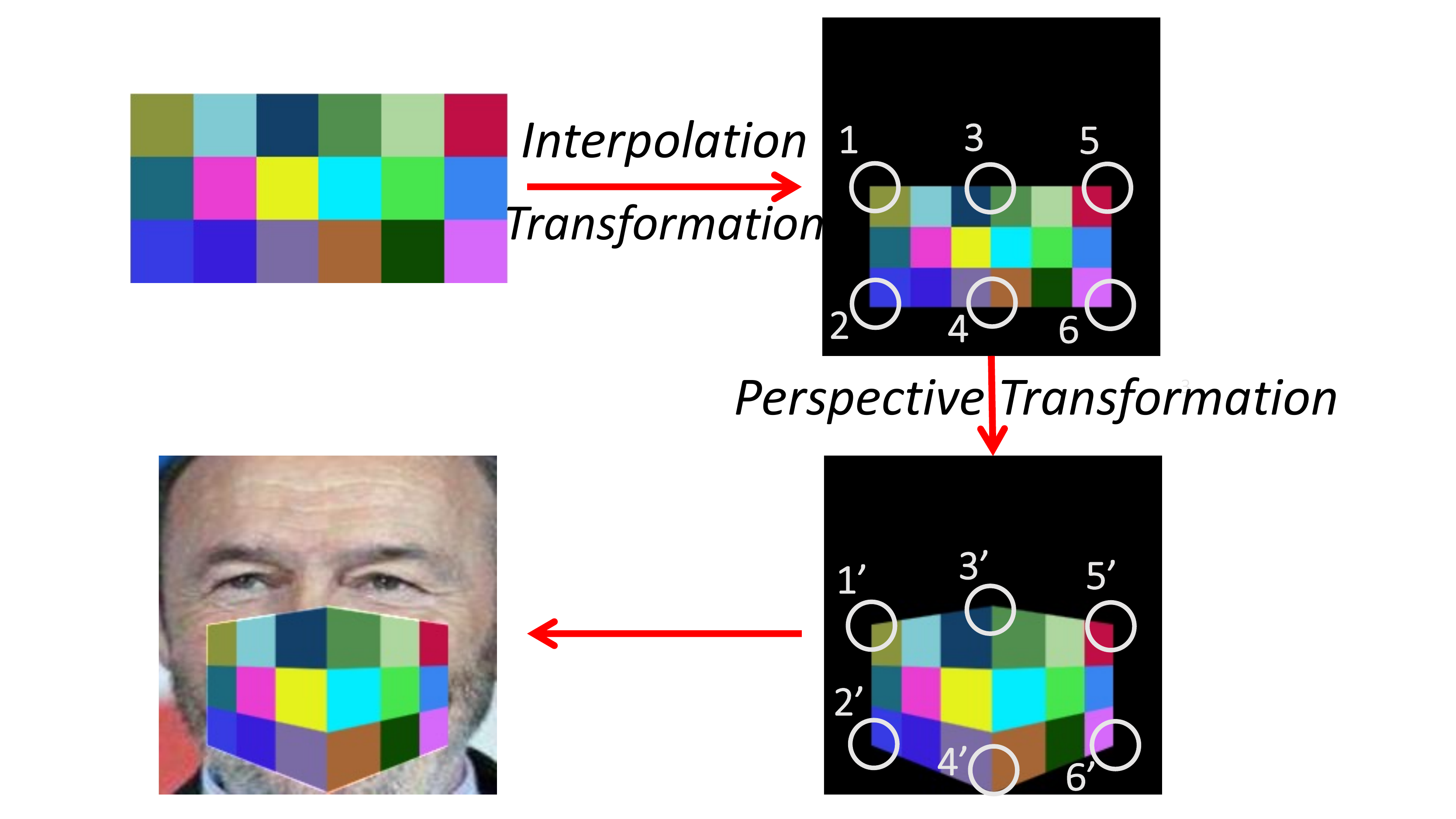} 
\caption{Transformations for the grid-level face mask attack.}
\label{fig:facemask}
\vspace{-8pt}
\end{figure}

\vspace{0.1cm}
\noindent{\bf Grid-level Physically Realizable Attack}.
In practice, pixel-level perturbations are not printable on face accessories made of \emph{coarse} materials, such as face masks using cloths and non-woven fabrics.
To address this issue, we propose the grid-level physically realizable face mask attack, which adds a color grid on face masks, as shown in Fig.~\ref{fig:perturbation_type}.  
Formally, the face mask attack on closed-set systems is formulated as the following optimization problem as a variation of Eq.~(\ref{eq:adv_example}) (formulations for other settings are presented in Appendix A):
\vspace{-5pt}
\begin{equation}
    \argmax_{\bm{\delta}} \ell (S(\bm{x}+M\cdot\mathcal{T}(\bm{\delta})), y),
\label{eq:facemask}
\vspace{-5pt}
\end{equation}
where $\bm{\delta}\in\mathbb{R}^{a\times b}$ is a $a\times b$ color matrix; each element of $\bm{\delta}$ represents an RGB color.
$M$ is the matrix that constrains the area of perturbations.
$\mathcal{T}$ is a sequence of transformations that convert $\bm{\delta}$ to a face mask with a color grid in digital space by the following steps, as shown in Fig.~\ref{fig:facemask}.
First, we use the \emph{interpolation transform} to scale up the color matrix $\bm{\delta}$ into a color grid in a background image, which has the same dimension as $\bm{x}$ and all pixel values set to be 0.
Then, we split the color grid into the left and right parts, each of which has four corner points.
Afterward, we use a \emph{perspective transformation} on each part of the grid for a 2-D alignment, which is based on the position of its source and destination corner points.
Finally, we add the aligned color grid onto the input face image $\bm{x}$.
Details of the perspective transformation and the algorithm for solving the optimization problem in Eq.~(\ref{eq:facemask}) can be found in Appendix A.

\subsection{Attacker's System Knowledge (K)}
\label{subsec:knowledge}

The key components of a face recognition system $S$ are the training set $D$ and neural architecture $h$.
It is natural to ask how do these two components contribute to the robustness against adversarial attacks.
From the attackers' perspective, we propose several evaluation scenarios in \model, which represent adversarial attacks performed under different knowledge levels on $D$ and $h$.

\vspace{0.1cm}

\noindent{\bf Zero Knowledge}.
Both $D$ and $h$ are invisible to the attacker, {\em i.e.}, $K = \emptyset$.
This is the weakest adversarial setting, as no critical information of $S$ is leaked. 
Thus, it provides an upper bound for robustness evaluation on $S$.
In this scenario, the attacks are referred to as \emph{black-box attacks}, where the attacker needs no internal details of $S$ to compromise it.



There are two general ways towards black-box attacks, \emph{query-based attack}~\cite{chen2017zoo, nitin2018practical} and \emph{transfer-based attack}~\cite{papernot2016transferability}. We employ the latter because the former attack requires a large number of online probes to repeatedly estimate the loss gradients of $S$ on adversarial examples, which is less practical than fully offline attacks when access to prediction decisions is unavailable. The latter method is built upon the \emph{transferability} of adversarial examples~\cite{papernot2016transferability, dong2018boosting}. Specifically, an attacker first collects a sufficient of training samples and builds a surrogate training set $D'$. Then, a surrogate system $S'$ is constructed by training a surrogate neural architecture $h'$ on $D'$ for the same task as $S$, \textit{i.e.}, $S'=f(h'; D')$.
Afterward, the attacker obtains a set of adversarial examples by performing \emph{white-box attacks} on the surrogate system $S'$, which constitutes the transferable adversarial examples for evaluating the robustness of $S$. 

\vspace{0.1cm}

\noindent{\bf Training Set}.
This scenario enables the assessment of the robustness of the training set of $S$ in adversarial settings.
Here, only the training set $D$ is visible to the attacker, \textit{i.e.}, $K = \{D\}$.
Without knowing $h$, an attacker constructs a surrogate system $S'$ by training a surrogate neural architecture $h'$ on $D$, \textit{i.e.,} $S'=f(h';D)$.
Then, the attacker performs the transfer-based attack aforementioned on $S'$ and evaluates $S$ by using the transferred adversarial examples. 

\vspace{0.1cm}

\noindent{\bf Neural Architecture}.
Similarly, the attacker may only know the neural architecture $h$ of $S$ but has no access to the training set $D$, \textit{i.e.}, $K=\{h\}$.
This enables us to evaluate the robustness of the neural architecture $h$ of $S$. 
Without knowing $D$, the attacker can build its surrogate system $S'=f(h;D')$ and conduct the transfer-based attack to evaluate $S$.  


\vspace{0.1cm}

\noindent{\bf Full Knowledge}.
In the worst case, the attacker can have an accurate knowledge of both the training set $D$ and neural architecture $h$ (\textit{i.e.}, $K=\{D, h\}$). Thus, it provides a lower bound for robustness evaluation on $S$. In this scenario, the attacker can fully reproduce $S$ in an offline setting and then performs \emph{white-box attacks} on $S$. 

The evaluation method described above is based on the assumption that the adversarial examples in response to a surrogate system $S'$ can always mislead the target system $S$. However, there is no theoretical guarantee, and recent studies show that some transferred adversarial examples can only fool the target system $S$ with a low success rate~\cite{liu2016delving}.

To boost the transferability of adversarial examples produced on the surrogate system, we leverage two techniques: \emph{momentum-based attack}~\cite{dong2018boosting} and \emph{ensemble-based attack}~\cite{liu2016delving, dong2018boosting}. 
First, inspired by the momentum-based attack, we integrate the \emph{momentum term} into the iterative process of the white-box attacks on the surrogate system $S'$ to stabilize the update directions and avoid the local optima. Thus, the resulting adversarial examples are more transferable.
Second, when the neural architecture $h$ of the target system $S$ is unavailable, we construct the surrogate system $S'$ using an ensemble of models with different neural architectures rather than a single model, \textit{i.e.}, $h'=\{h'_i\}_{i=1}^k$, where $\{h'_i\}_{i=1}^k$ is an ensemble of $k$ models.
Specifically, we aggregate the output logits of $h_i(i \leq k)$ in a similar way to \cite{dong2018boosting}.
The rationale behind this is that if an adversarial example can fool multiple models, it is more likely to mislead other models.

\begin{table*}[]
\vspace{-18pt}
\centering
\caption{Optimization formulations by the attacker's goal.}
\scalebox{0.90}{
\begin{tabular}{|c|c|c|}
\hline
\textbf{Target System} & \textbf{Attacker's Goal} & \textbf{Formulation} \\ \hline\hline
Closed-set    & Dodging         & $\max_{\bm{\delta}} \ell(S(\bm{x}+M\bm{\delta}), y), \ \ \ \ s.t.\  ||\bm{\delta}||_p \leq \epsilon$  \\ \hline
Closed-set    & Impersonation   & $\min_{\bm{\delta}} \ell(S(\bm{x}+M\bm{\delta}), y_t), \ \ \ \ s.t.\  ||\bm{\delta}||_p \leq \epsilon$  \\ \hline
Open-set      & Dodging         & $\max_{\bm{\delta}} d(S(\bm{x}+M\bm{\delta}), S(\bm{x}^{*})), \ \ \ \ s.t.\  ||\bm{\delta}||_p \leq \epsilon$              \\ \hline
Open-set      & Impersonation   & $\min_{\bm{\delta}} d(S(\bm{x}+M\bm{\delta}), S(\bm{x}^{*}_t)), \ \ \ \ s.t.\  ||\bm{\delta}||_p \leq \epsilon$              \\ \hline
\end{tabular}
}
\label{tab:objective}
\vspace{-5pt}
\end{table*}

\subsection{Attacker's Goal (G)} 

In addition to the attacker's system knowledge about $S$, adversarial attacks can differ in specific goals. In \model, we are interested in the following two types of attacks with different goals:

\vspace{0.1cm}

\noindent{\bf Dodging/Non-targeted}.
In a dodging attack, an attacker aims to have his/her face misidentified as another arbitrary face. 
\textit{e.g.}, the attacker can be a terrorist who wants to bypass a face recognition system for biometric security checking.
As the dodging attack has no specific identity as which it aims to predict an input face image, it is also called the \emph{non-targeted attack}.

\vspace{0.1cm}

\noindent{\bf Impersonation/Targeted}.
In an impersonation/targeted attack, an attacker seeks to produce an adversarial example that is misrecognized as a target identity. For example, the attacker may try to camouflage his/her face to be identified as an authorized user of a laptop, which uses face recognition for authentication.

In \model, we formulate the dodging attack and impersonation attack as constrained optimization problems, corresponding to different face recognition systems and the attacker's goals, as shown in Table~\ref{tab:objective}.
Here, $\ell$ denotes the softmax cross-entropy loss used in closed-set systems, $d$ represents the distance metric for open-set systems (\textit{e.g.}, the cosine distance obtained by subtracting cosine similarity from one), $(\bm{x}, y)$ is the input face image and the associated identity, $\bm{\delta}$ is the adversarial perturbation, $S$ represents a face recognition system which is built on either a single model or an ensemble of models with different neural architectures, $M$ denotes the mask matrix that constrains the area of perturbation (similar to Eq.~(\ref{eq:adv_example})), $\epsilon$ is the $\ell_p$-norm bound of $\bm{\delta}$.
For closed-set systems, we use $y_t$ to represent the target identity of impersonation attacks.
For open-set systems, we use $\bm{x}^*$ to denote the gallery face image that belongs to the identity as $\bm{x}$, and $\bm{x}_t^{*}$ as the gallery image for the target identity of impersonation.

Note that the formulations listed in Table~\ref{tab:objective} work for both digital attacks and physically realizable attacks:
For the former, we use a small value of $\epsilon$ 
and let $M$ be an all-one matrix to ensure imperceptible perturbations on the entire image.
For the latter, we use a large $\epsilon$ 
and let $M$ to constrain $\bm{\delta}$ in a small area of $\bm{x}$. 



\subsection{Attacker's Capability (C)} 
\label{subsec:capability}


In practice, even when the attackers share the same system knowledge and goal, their capabilities can still be different due to the time and/or budget constraints, such as the budget for printing adversarial eyeglass frames~\cite{sharif2016accessorize}.
Thus, in \model, we consider two types of attacks corresponding to different attacker's capabilities: \emph{individual attack} and \emph{universal attack}.

\vspace{0.1cm}
\noindent{\bf Individual Attack}.
The attacker has a strong capability with enough time and budget to produce a specific perturbation for each input face image.
In this case, the optimization formulations are the same as those shown in Table~\ref{tab:objective}.

\vspace{0.1cm}
\noindent{\bf Universal Attack}.
The attacker has a time/budget constraint such that he/she is only able to generate a \emph{face-agnostic} perturbation that fools a face recognition system on a batch of face images instead of every input.

One common way to compute a universal perturbation is to sequentially find the \emph{minimum} perturbation of each data point in the batch and then aggregate these perturbations~\cite{moosavi2017universal}.
However, this method requires orders of magnitude running time: it processes only one image at each iteration, so a large number of iterations are needed to obtain a satisfactory universal perturbation. 
Moreover, it only focuses on digital attacks and cannot be generalized to physically realizable attacks, which seek large perturbations in a restricted area rather than the minimum perturbations. 

To address these issues, we formulate the universal attack as a \emph{maxmin optimization} as follows (using the dodging attack on closed-set systems as an example):
\begin{equation}
\vspace{-2pt}
   \max_{\bm{\delta}}\min\{ \ell(S(\bm{x}_i+M\bm{\delta}), y_i) \}_{i=1}^{N}, \ \  s.t.\  ||\bm{\delta}||_p \leq \epsilon ,
\label{eq:universal}
\vspace{-2pt}
\end{equation}
where $\{\bm{x}_i, y_i\}_{i=1}^N$ is a batch of input images that share the universal perturbation $\bm{\delta}$.
Compared to ~\cite{moosavi2017universal}, our approach has several advantages:
First, we can significantly improve the efficiency by processing images batchwise.
Second, our formulation can explicitly control the universality of the perturbation by setting different values of $N$.
Third, our method can be generalized to both digital attacks and physically realizable attacks.
Details of our algorithm for solving the optimization problem in Eq.~(\ref{eq:universal}) and the formulations for other settings can be found in Appendix B. 


\section{Experiments}
\label{sec:experiments}

In this section, we evaluate a variety of face recognition systems using \model\ on both closed-set and open-set tasks under different adversarial settings. 

\subsection{Experimental Setup}

\begin{table}[]
\centering
\vspace{-5pt}
\caption{Open-set face recognition systems in our experiments.}
\scalebox{0.7}{
\begin{tabular}{|c|c|c|c|}
\hline
\textbf{Target Model}      & \textbf{Training Set}  & \textbf{Neural Architecture} & \textbf{Loss}    \\ \hline
VGGFace~\cite{parkhi2015deep}    & VGGFace~\cite{parkhi2015deep}       & VGGFace~\cite{parkhi2015deep}             & Triplet~\cite{parkhi2015deep} \\ \hline
FaceNet~\cite{facenetpytorch}    & CASIA-WebFace~\cite{yi2014learning} & InceptionResNet~\cite{szegedy2016inception}     & Triplet~\cite{schroff2015facenet} \\ \hline
ArcFace18~\cite{arcfacepytorch}  & MS-Celeb-1M~\cite{guo2016ms}   & IResNet18~\cite{Liang2018Learning}           & ArcFace~\cite{deng2019arcface} \\ \hline
ArcFace50~\cite{arcfacepytorch}  & MS-Celeb-1M~\cite{guo2016ms}   & IResNet50~\cite{Liang2018Learning}           & ArcFace~\cite{deng2019arcface} \\ \hline
ArcFace101~\cite{arcfacepytorch} & MS-Celeb-1M~\cite{guo2016ms}   & IResNet101~\cite{Liang2018Learning}          & ArcFace~\cite{deng2019arcface} \\ \hline
\end{tabular}
}
\label{tab:open-set}
\vspace{-8pt}
\end{table}

{\bf Datasets}.
For closed-set systems, we use a subset of the VGGFace2 dataset~\cite{cao2018vggface2}.
Specifically, we select 100 classes, each of which has 181 face images.
For open-set systems, we employ the VGGFace2, MS-Celeb-1M~\cite{guo2016ms}, CASIA-WebFace~\cite{yi2014learning} datasets for training surrogate models, and the LFW dataset~\cite{LFWTech} for testing.

{\bf Neural Architectures}.
The face recognition systems with five different neural networks are evaluated in our experiments: VGGFace~\cite{parkhi2015deep}, InceptionResNet~\cite{szegedy2016inception}, IResNet18~\cite{Liang2018Learning}, IResNet50~\cite{Liang2018Learning}, and IResNet101~\cite{Liang2018Learning}.



{\bf Evaluation Metric}.
We use \emph{attack success rate} = 1 - accuracy as the evaluation metric.
Specifically, a higher attack success rate indicates that a face recognition system is more fragile in adversarial settings, while a lower rate shows higher robustness against adversarial attacks. 

{\bf Implementation}.
For open-set face recognition, we directly applied five publicly available pre-trained face recognition models 
as the target models for attacks, as summarized in Table~\ref{tab:open-set}.
At prediction stage, we used $100$ photos randomly selected from frontal images in the LFW dataset~\cite{LFWTech}, each of which is aligned by using MTCNN~\cite{zhang2016joint} and corresponds to one identity.
And we used another $100$ photos of the same identities as the test gallery.
We computed the cosine similarity between the feature vectors of the test and gallery photos. If the score is above a threshold corresponding to a False Acceptance Rate of $0.001$, then the test photo is predicted to have the same identity as the gallery photo.

For closed-set face recognition, we randomly split each class of the VGGFace2 subset into three parts: $150$ for training, $30$ for validation, and $1$ for testing.  
To train closed-set models, we used standard transfer learning with the open-set models listed in Table~\ref{tab:open-set}.
Specifically, we initialized each closed-set model with the corresponding open-set model, and then added a final fully connected layer, which contains $100$ neurons.
Unless otherwise specified, each model was trained for $60$ epochs with a training batch size of $64$. We used the Adam optimizer~\cite{kingma2015AdamAM} with an initial learning rate of $0.0001$, then dropped the learning rate by $0.1$ at the 20th and 35th epochs.

\begin{figure}
\vspace{-5pt}
\centering
 \includegraphics[width=0.45\textwidth]{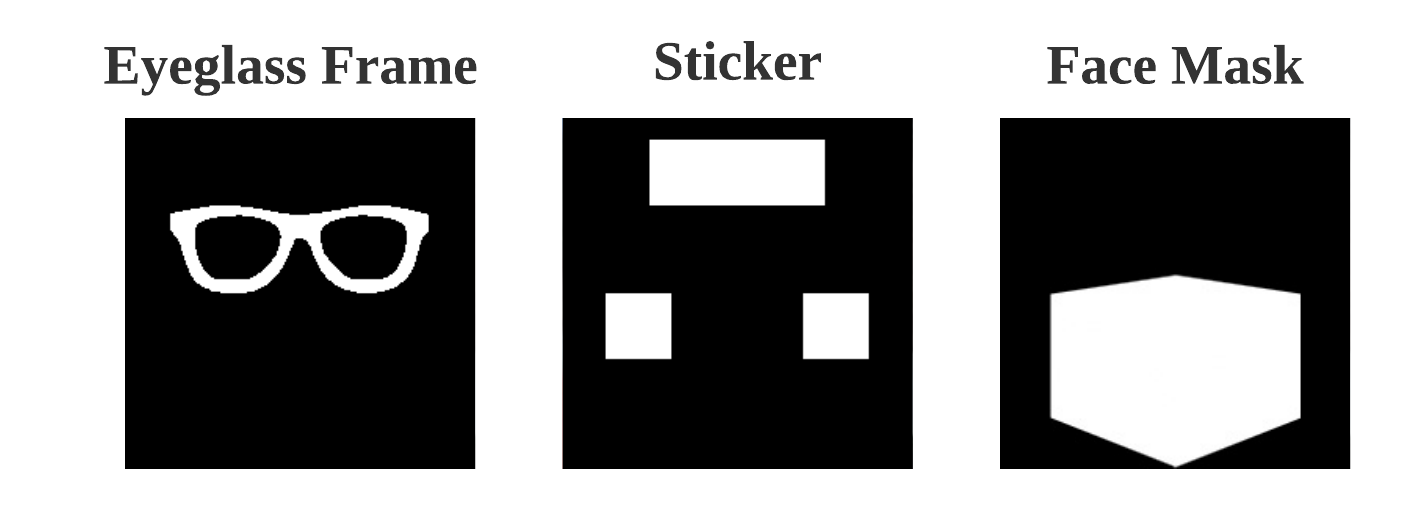} 
\caption{Mask matrices for physically realizable attacks in \model.}
\label{fig:mask_matrix}
\vspace{-8pt}
\end{figure}

For each physically realizable attack in \model, we used $255/255$ as the $\ell_\infty$ norm bound for perturbations allowed, and ran each attack for $200$ iterations.
For the PGD attack~\cite{madry2018towards}, we used an $\ell_\infty$ bound $8/255$ and $40$ iterations. The dimension of the color grid for face mask attacks is set to $16\times 8$. 
The mask matrices that constrain the areas of perturbations for physically realizable attacks are visualized in Fig.~\ref{fig:mask_matrix}.

\subsection{Robustness of Face Recognition Components}
We begin by using \model~to assess the robustness of face recognition components in various adversarial settings.
For a given target face recognition system $S$ and a perturbation type $P$, we evaluate the training set $D$ and neural architecture $h$ of $S$ with the four evaluation scenarios presented in Section~\ref{subsec:knowledge}.
Specifically, when $h$ is invisible to the attacker, we construct the surrogate system $S'$ by ensembling the models built on the other four neural architectures shown in Table~\ref{tab:open-set}.
In the scenarios where the attacker has no access to $D$, we build the surrogate training set $D'$ with another VGGFace2 subset that has the same classes as $D$ in closed-set settings, and use the other four training sets listed in Table~\ref{tab:open-set} for open-set tasks.
We present the experimental results for dodging attacks on closed-set face recognition systems in Table \ref{tab:attack-closed}, and the results for zero-knowledge dodging attacks on open-set VGGFace and FaceNet in Table \ref{tab:attack-open}. 
The other results can be found in Appendix C. 
Additionally, we evaluate the efficacy of using \emph{momentum} and \emph{ensemble} methods to improve transferability of adversarial examples, which is detailed in Appendix D.

\begin{table}[t]
\centering
\vspace{-5pt}
\caption{Attack success rate of dodging attacks on closed-set face recognition systems by the attacker's system knowledge. Z represents zero knowledge, T is training set, A is neural architecture, and F represents full knowledge.}
\scalebox{0.80}{
\begin{tabular}{|c|c|c|c|c|c|}
\hline
\multirow{2}{*}{\textbf{Target System}} & \multirow{2}{*}{\textbf{Attack Type}} & \multicolumn{4}{c|}{\textbf{Attacker's System Knowledge}} \\ \cline{3-6} 
                                        &                                       & \textbf{Z}   & \textbf{T}   & \textbf{A}   & \textbf{F}   \\ \hline
\multirow{4}{*}{VGGFace}                & PGD                                   & 0.40         & 0.51         & 0.93         & 0.94         \\ \cline{2-6} 
                                        & Eyeglass Frame                        & 0.23         & 0.28         & 0.70         & 0.99         \\ \cline{2-6} 
                                        & Sticker                               & 0.05         & 0.06         & 0.47         & 0.98         \\ \cline{2-6} 
                                        & Face Mask                             & 0.26         & 0.32         & 0.63         & 1.00         \\ \hline\hline
\multirow{4}{*}{FaceNet}                & PGD                                   & 0.83         & 0.83         & 1.00         & 1.00         \\ \cline{2-6} 
                                        & Eyeglass Frame                        & 0.13         & 0.16         & 0.90         & 1.00         \\ \cline{2-6} 
                                        & Sticker                               & 0.01         & 0.01         & 0.92         & 1.00         \\ \cline{2-6} 
                                        & Face Mask                             & 0.30         & 0.42         & 0.83         & 1.00         \\ \hline\hline
\multirow{4}{*}{ArcFace18}              & PGD                                   & 0.87         & 0.92         & 0.97         & 1.00         \\ \cline{2-6} 
                                        & Eyeglass Frame                        & 0.06         & 0.06         & 0.44         & 1.00         \\ \cline{2-6} 
                                        & Sticker                               & 0.01         & 0.01         & 0.37         & 1.00         \\ \cline{2-6} 
                                        & Face Mask                              & 0.27         & 0.33         & 0.71         & 1.00         \\ \hline\hline
\multirow{4}{*}{ArcFace50}              & PGD                                   & 0.87         & 0.90         & 0.81         & 0.99         \\ \cline{2-6} 
                                        & Eyeglass Frame                        & 0.09         & 0.12         & 0.44         & 0.99         \\ \cline{2-6} 
                                        & Sticker                               & 0.00         & 0.01         & 0.14         & 0.94         \\ \cline{2-6} 
                                        & Face Mask                             & 0.29         & 0.36         & 0.67         & 0.99         \\ \hline\hline
\multirow{4}{*}{ArcFace101}             & PGD                                   & 0.81         & 0.78         & 0.86         & 0.96         \\ \cline{2-6} 
                                        & Eyeglass Frame                        & 0.03         & 0.03         & 0.26         & 0.98         \\ \cline{2-6} 
                                        & Sticker                               & 0.04         & 0.04         & 0.08         & 0.95         \\ \cline{2-6} 
                                        & Face Mask                             & 0.26         & 0.36         & 0.54         & 0.99         \\ \hline
\end{tabular}
}
\label{tab:attack-closed}
\vspace{-5pt}
\end{table}

\begin{table}[t]
\centering
\caption{Attack success rate of dodging attacks on open-set face recognition systems with zero knowledge.}
\scalebox{0.80}{
\begin{tabular}{|c|c|c|c|c|}
\hline
\multirow{2}{*}{\textbf{Target Model}} & \multicolumn{4}{c|}{\textbf{Attack Type}}          \\ \cline{2-5} 
                              & \textbf{PGD}  & \textbf{Sticker} & \textbf{Eyeglass Frame} & \textbf{Face Mask} \\ \hline\hline
VGGFace                       & 0.26 & 0.56    & 0.79     & 0.67      \\ \hline
FaceNet                       & 0.55 & 0.13    & 0.54     & 0.62      \\ \hline
\end{tabular}
}
\label{tab:attack-open}
\vspace{-8pt}
\end{table}

It can be seen from Table \ref{tab:attack-closed} that: \emph{the neural architecture is significantly more fragile than the training set in most adversarial settings}.
For example, when only the neural architecture is exposed to the attacker, the sticker attack has a high success rate of $0.92$ on FaceNet. In contrast, when the attacker only knows the training set, the attack success rate significantly drops to $0.01$. In addition, by comparing each row of Table \ref{tab:attack-closed} that corresponds to the same target system, we observe that \emph{digital attacks (PGD) are considerably more potent than their physically realizable counterparts on closed-set systems, while grid-level perturbations on face masks are noticeably more effective than pixel-level physically realizable perturbations (\textit{i.e.}, the eyeglass frame attack and the sticker attack).}
Moreover, by comparing the zero knowledge attacks in Table~\ref{tab:attack-closed} and \ref{tab:attack-open}, we find that \emph{open-set face recognition systems are more vulnerable than closed-set systems} such that nearly all perturbation types of attacks (even the black-box sticker attack that often fails in closed-set) 
tend to be more likely to successfully transfer
across different open-set systems (\textit{i.e.}, 
these are more susceptible to black-box attacks), which should raise more concerns about their security.


\subsection{Robustness Under Universal Attacks}

\begin{table}[]
\centering
\vspace{-15pt}
\caption{Attack success rate of dodging attacks on closed-set face recognition systems by the universality of adversarial examples.
Here, $N$ represents the batch size of face images that share a universal perturbation. 
}
\scalebox{0.80}{
\begin{tabular}{|c|c|c|c|c|c|}
\hline
\multirow{2}{*}{\textbf{Target System}} & \multirow{2}{*}{\textbf{Attack Type}} & \multicolumn{4}{c|}{\textbf{Attacker's Capability}} \\ \cline{3-6} 
                                        &                                       & \textbf{N=1}   & \textbf{N=5}   & \textbf{N=10}   & \textbf{N=20}   \\ \hline
\multirow{4}{*}{VGGFace}                & PGD                                   & 0.94   & 0.86   & 0.31   & 0.15         \\ \cline{2-6} 
                                        & Eyeglass Frame                        & 0.99   & 0.91   & 0.52   & 0.23         \\ \cline{2-6} 
                                        & Sticker                               & 0.98   & 0.66   & 0.34   & 0.09         \\ \cline{2-6} 
                                        & Face Mask                             & 1.00   & 1.00   & 0.88   & 0.56         \\ \hline\hline
\multirow{4}{*}{FaceNet}                & PGD                                   & 1.00   & 1.00   & 0.80   & 0.21         \\ \cline{2-6} 
                                        & Eyeglass Frame                        & 1.00   & 1.00   & 1.00   & 0.62         \\ \cline{2-6} 
                                        & Sticker                               & 1.00   & 1.00   & 0.98   & 0.61         \\ \cline{2-6} 
                                        & Face Mask                             & 1.00   & 1.00   & 1.00   & 0.91          \\ \hline\hline
\multirow{4}{*}{ArcFace18}              & PGD                                   & 1.00   & 1.00   & 0.64   & 0.08         \\ \cline{2-6} 
                                        & Eyeglass Frame                        & 1.00   & 0.96   & 0.44   & 0.08         \\ \cline{2-6} 
                                        & Sticker                               & 1.00   & 0.56   & 0.09   & 0.00         \\ \cline{2-6} 
                                        & Face Mask                              & 0.99   & 0.92   & 0.90   & 0.67         \\ \hline\hline
\multirow{4}{*}{ArcFace50}              & PGD                                   & 1.00   & 0.80   & 0.37   & 0.05         \\ \cline{2-6} 
                                        & Eyeglass Frame                        & 0.99   & 0.81   & 0.38   & 0.07         \\ \cline{2-6} 
                                        & Sticker                               & 0.91   & 0.28   & 0.06   & 0.00         \\ \cline{2-6} 
                                        & Face Mask                             & 0.99   & 0.98   & 0.81   & 0.72         \\ \hline\hline
\multirow{4}{*}{ArcFace101}             & PGD                                  & 0.96   & 0.91   & 0.24   & 0.03         \\ \cline{2-6} 
                                        & Eyeglass Frame                       & 0.98   & 0.71   & 0.19   & 0.02         \\ \cline{2-6} 
                                        & Sticker                               & 0.93   & 0.15   & 0.03   & 0.00         \\ \cline{2-6} 
                                        & Face Mask                             & 0.99   & 0.92   & 0.90   & 0.67         \\ \hline
\end{tabular}
}
\label{tab:universal}
\vspace{-9pt}
\end{table}


Next, we use \model\ to evaluate the robustness of face recognition systems with various extents of adversarial universality by setting the parameter $N$ in Eq.~(\ref{eq:universal}) to different values. For a given $N$, we split the testing set into mini-batches of size $N$, and produce a specific perturbation for each batch. Note that when $N=1$, a universal attack is reduced to an individual attack. Table~\ref{tab:universal} shows the experimental results for universal dodging attacks on closed-set systems. The other results are presented in Appendix E.


Our first observation is that \emph{face recognition systems are significantly more vulnerable to the universal face masks than other types of universal perturbations.}
Under a large extent of universality (\textit{e.g.}, when $N=20$), face mask attacks remain $>0.5$ success rates.
Particularly noteworthy is the universal face mask attacks on FaceNet, which can achieve a rate as high as $0.91$. 
In contrast, other universal attacks can have relatively low success rates (\textit{e.g.}, $0.08$ for eyeglass frame attack on ArcFace18).
The second observation is that \emph{the robustness of a face recognition system against different types of universal perturbations is highly dependent on its neural architecture}. For example, the ArcFace101 architecture is more robust than the others in most settings, while FaceNet tends to be the most fragile one.

\subsection{Is ``Robust'' Face Recognition Really Robust?}

\begin{figure}
\centering
\vspace{-15pt}
 \includegraphics[width=0.45\textwidth]{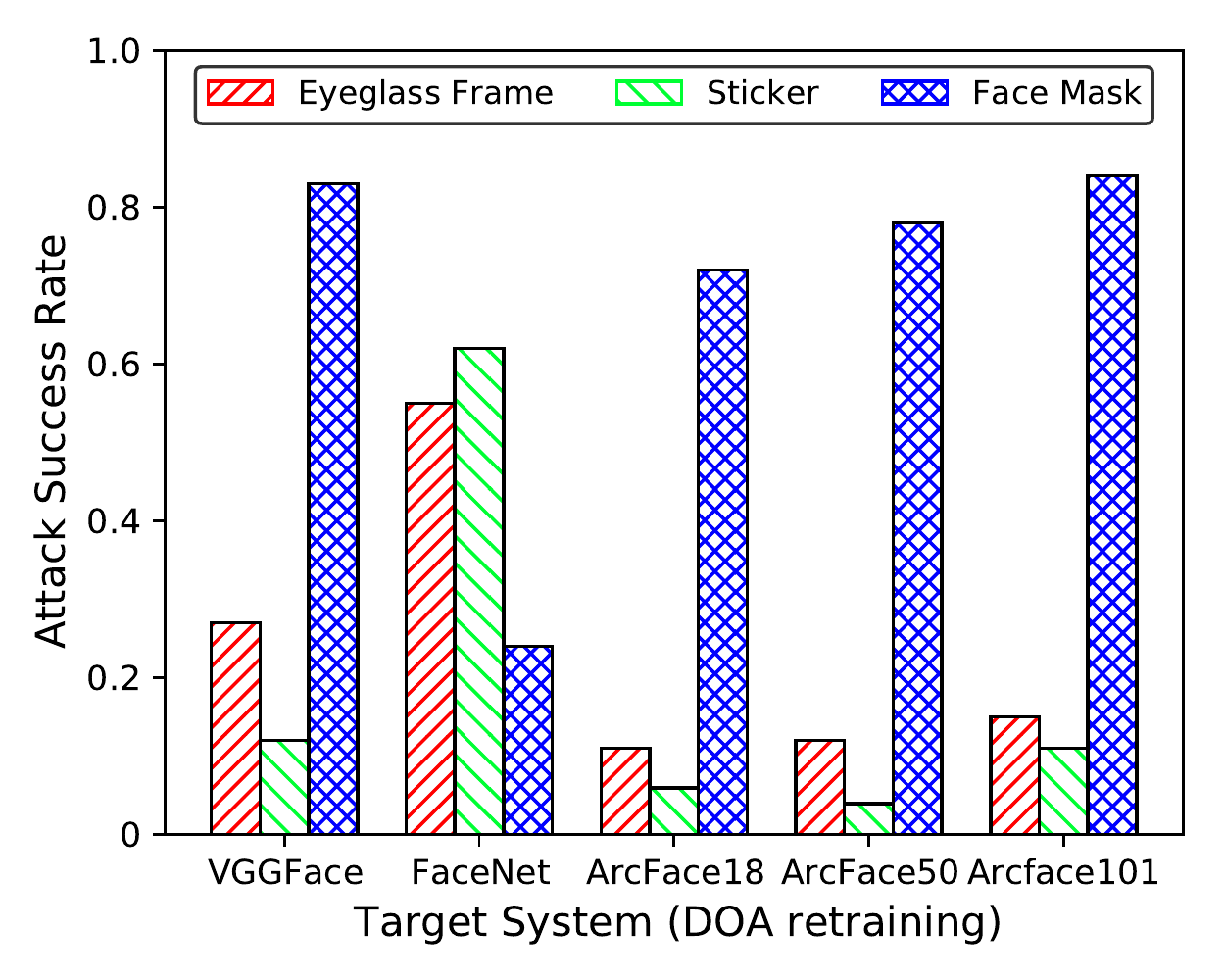} 
\caption{Attack success rate of dodging physically realizable attacks on closed-set systems with DOA retraining.}
\label{fig:DOA}
\vspace{-15pt}
\end{figure}

While numerous approaches have been proposed for making deep neural networks more robust to adversarial examples, only a few~\cite{wu2019defending} focus on defending against physically realizable attacks on face recognition systems. These defense approaches have achieved good performance for certain types of realizable attacks and neural architectures, but their effectiveness for other types of attacks and face recognition systems remains unknown.
In this section, we apply \model\ to evaluate the state-of-the-art defense paradigms.
Specifically, we first use DOA~\cite{wu2019defending}, a method that defends closed-set VGGFace against eyeglass frame attacks~\cite{sharif2016accessorize} to retrain each closed-set system. We then evaluate the refined systems using the three physically realizable attacks included in \model.  
Fig.~\ref{fig:DOA} shows the experimental results for dodging attacks.

Our first observation is that \emph{the state-of-the-art defense approach, DOA, fails to defend against the grid-level perturbations on face masks for most neural architectures}. Specifically, face mask attacks can achieve $>0.7$ success rates on four out of the five face recognition systems refined by DOA. Moreover, we observe that \emph{adversarial robustness against one type of perturbation can not be generalized to other types}.
For example, while VGGface-DOA exhibits a relatively high level of robustness (more than a $70\%$ accuracy) against pixel-level perturbations (\textit{i.e.}, stickers and eyeglass frames), it is very vulnerable to grid-level perturbations (\textit{i.e.}, face masks). In contrast, using DOA on FaceNet can successfully defend face mask perturbations with the attack success rate significantly dropping from $1.0$ to $0.24$, but it's considerably less effective against eyeglass frames and stickers. In summary, these results show that the effectiveness of defense is highly dependent on the nature of perturbation and neural architectures, which in turn, indicates that it is critical to consider different types of attacks and neural architectures when evaluating a defense method for face recognition systems.

\vspace{-5pt}
\section{Conclusion}
\label{sec:conclusion}

We present \model, a fine-grained robustness evaluation framework for face recognition systems.
\model\ incorporates four evaluation dimensions and can work on both face identification and verification of open-set and closed-set systems.
To our best knowledge, \model\ is the first-of-its-kind platform that supports to evaluate the risks of different components of face recognition systems from multiple dimensions and under various adversarial settings.
The comprehensive and systematic evaluations on five state-of-the-art face recognition systems demonstrate that \model\ can greatly help understand the robustness of the systems against both digital and physically realizable attacks.
We envision that \model\ can serve as a useful framework to advance future research of adversarial learning on face recognition.

\balance
\bibliographystyle{ieee_fullname}
\bibliography{facesec}

\newpage
\appendix
\nobalance
\begin{table*}[htp]
\centering
\caption{Optimization formulations of grid-level face mask attacks.}
\scalebox{1.0}{
\begin{tabular}{|c|c|c|}
\hline
\textbf{Target System} & \textbf{Attacker's Goal} & \textbf{Formulation} \\ \hline\hline
Closed-set    & Dodging         & $\max_{\bm{\delta}} \ell(S(\bm{x}+M\cdot\mathcal{T}(\bm{\delta})), y)$  \\ \hline
Closed-set    & Impersonation   & $\min_{\bm{\delta}} \ell(S(\bm{x}+M\cdot\mathcal{T}(\bm{\delta})), y_t)$  \\ \hline
Open-set      & Dodging         & $\max_{\bm{\delta}} d(S(\bm{x}+M\cdot\mathcal{T}(\bm{\delta})), S(\bm{x}^{*}))$              \\ \hline
Open-set      & Impersonation   & $\min_{\bm{\delta}} d(S(\bm{x}+M\cdot\mathcal{T}(\bm{\delta})), S(\bm{x}^{*}_t))$              \\ \hline
\end{tabular}
}
\label{tab:face_mask}
\end{table*}

\begin{table*}[t]
\centering
\caption{Optimization formulations of universal dodging attacks.}
\scalebox{1.00}{
\begin{tabular}{|c|c|c|}
\hline
\textbf{Target System} & \textbf{Perturbation Type} & \textbf{Formulation} \\ \hline\hline
Closed-set    & Pixel-level         & $\max_{\bm{\delta}}\min\{ \ell(S(\bm{x}_i+M\bm{\delta}), y_i) \}_{i=1}^{N}, \ \  s.t.\  ||\bm{\delta}||_p \leq \epsilon$  \\ \hline
Closed-set    & Grid-level   & $\max_{\bm{\delta}}\min\{ \ell(S(\bm{x}_i+M\cdot\mathcal{T}(\bm{\delta})), y_i) \}_{i=1}^{N}$  \\ \hline
Open-set      & Pixel-level         & $\max_{\bm{\delta}} \min \{ d(S(\bm{x_i}+M\bm{\delta}), S(\bm{x}_i^{*})) \}_{i=1}^{N}, \ \ \ \ s.t.\  ||\bm{\delta}||_p \leq \epsilon$              \\ \hline
Open-set      & Grid-level   & $\max_{\bm{\delta}} \min \{ d(S(\bm{x_i}+M\cdot\mathcal{T}(\bm{\delta})), S(\bm{x}_i^{*})) \}_{i=1}^{N}$  \\ \hline
\end{tabular}
}
\label{tab:universal}
\end{table*}

\section{Grid-level Face Mask Attack}
\subsection{Formulation}
The optimization formulations of the proposed grid-level face mask attacks under different settings are presented in Table~\ref{tab:face_mask}.
Here, $S$ is the target face recognition model, $\bm{x}$ is the original input face image.
$\bm{\delta}\in\mathbb{R}^{a\times b}$ is a $a\times b$ color matrix; each element of $\bm{\delta}$ represents an RGB color.
$M$ denotes the mask matrix that constrains the area of perturbation; it contains $1$s where perturbation is allowed, and $0$s where there is no perturbation. 
For closed-set systems, $\ell$ denotes the softmax cross-entropy loss function, $y$ is the identity of $\bm{x}$, and $y_t$ is the target identity for impersonation attacks.
For open-set settings, $d$ is the cosine distance (obtained by subtracting cosine similarity from one), $\bm{x}^*$ is the gallery image of $\bm{x}$, and $\bm{x}^*_t$ is the target gallery image for impersonation. 
$\mathcal{T}$ represents a set of transformations that convert the color matrix $\bm{\delta}$ to a face mask with a color grid in digital space.
Specifically, $\mathcal{T}$ contains two transformations: \emph{interpolation transformation} and \emph{perspective transformation}, which are detailed below.

\subsection{Interpolation Transformation}

The interpolation transform starts from a $a \times b$ color matrix $\bm{\delta}$ and uses the following two steps to scale $\bm{\delta}$ into a face image, as illustrated in Fig.~\ref{fig:facemask}:
First, it resizes the color matrix from $a \times b$ to a rectangle $\bm{\delta}'$ with $c \times d$ pixels, so as to reflect the size of a face mask in a face image in digital space while preserving the layout of the color grids represented by $\bm{\delta}$.  
Specifically, in \model, each input face image has $224\times 224$ pixels. Let $(a,b)=(8,16)$ and $(c, d)= (80, 160)$.
Then, we put the face mask $\bm{\delta}'$ into a background image, such that the pixels in the rectangular area have the same value with $\bm{\delta}'$, and those outside the face mask area have values of $0$s.  

\begin{figure}[t]
\centering
 \includegraphics[width=0.4\textwidth]{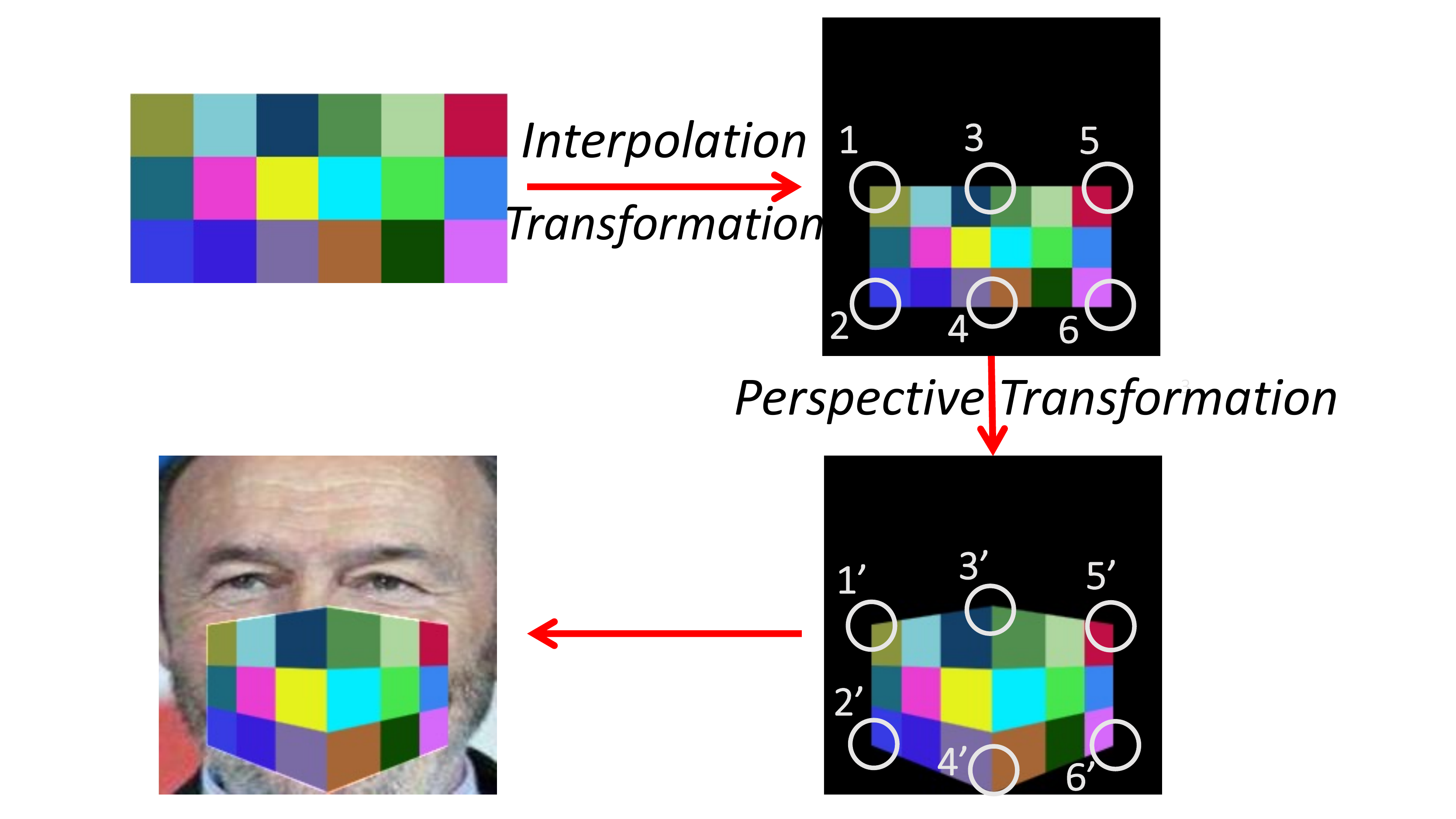} 
\caption{Transformations for the grid-level face mask attack.}
\label{fig:facemask}
\end{figure}

\subsection{Perspective Transformation}

Once the rectangle $\bm{\delta}'$ is embedded into a background image, we use a 2-D alignment that relies on the perspective transformation by the following steps.
First, we divide $\bm{\delta}'$ into a left half part $\bm{\delta}'_L$ and a right half part $\bm{\delta}'_R$; each is rectangular with four corners.
Then, we apply the perspective transformation to project each part to be with aligned coordinates, such that the new coordinates align with the position when a face mask is put on a human face, as shown in  Fig.~\ref{fig:facemask}.
Let $\bm{\delta}^{''}_L$ and $\bm{\delta}^{''}_R$ be the left and right part of the aligned face mask, the perspective transformation aims to find a $3\times 3$ matrix $N_k$ ($k\in\{L, R\}$) for each part such that the coordinates satisfy:
\[ \bm{\delta}^{''}_k (x, y) = \bm{\delta}'_k (u, v), \ \  k\in\{L, R\}, \]
where 
\[ u = \frac{N_k(1,1) x + N_k(1,2) y + N_k(1,3)}{N_k(3,1) x + N_k(3,2) y + N_k(3,3)}, \]
and 
\[ v = \frac{N_k(2,1) x + N_k(2,2) y + N_k(2,3)}{N_k(3, 1) x + N_k(3,2) y + N_k(3, 3)}.\]
Finally, we merge $\bm{\delta}^{''}_L$ and $\bm{\delta}^{''}_R$ to obtain the aligned grid-level face mask.

\renewcommand{\algorithmicrequire}{ \textbf{Input:}} 
\renewcommand{\algorithmicensure}{ \textbf{Output:}} 

\begin{algorithm}[t]     
\begin{algorithmic}[1]  
\REQUIRE 

Target system $S$;\\
Input face image $\bm{x}$ and its identity $y$; \\
The number of iterations $T$; \\
Step size $\alpha$; \\
Momentum parameter $\mu$. 
\ENSURE  The color matrix of adversarial face mask $\bm{\delta}_T$.
\STATE Initialize the color matrix $\bm{\delta}_0:= \bm{0}$, momentum $\bm{g}_0:= \bm{0}$;
\STATE Use interpolation and perspective transformations to convert $\bm{\delta}_0$:
$\bm{\delta}^{''}_0:= \mathcal{T}(\bm{\delta}_0)$;
\FOR{each $t \in [0, T-1]$}
\STATE $\bm{g}_{t+1}:=\mu \cdot \bm{g}_t + \frac{\nabla_{\bm{\delta}_t}\ell(S(\bm{x}+M\cdot\bm{\delta}^{''}_t), y)}{||\ell(S(\bm{x}+M\cdot\bm{\delta}^{''}_t), y)||_1}$;
\STATE $\bm{\delta}_{t+1}:= \bm{\delta}_t + \alpha\cdot\mathrm{sign}(\bm{g}_{t+1})$;
\STATE $\bm{\delta}^{''}_{t+1} := \mathcal{T}(\bm{\delta}_{t+1})$;
\STATE Clip $\bm{\delta}^{''}_{t+1}$ such that pixel values of $\bm{x}+M\cdot\bm{\delta}^{''}_{t+1}$ are in $[0, 255/255]$;
\ENDFOR  
\RETURN $\bm{\delta}_T$.  
\end{algorithmic}
\caption{Computing adversarial face mask.}
\label{alg:face_mask} 
\end{algorithm}

\subsection{Computing Adversarial Face Masks}
The algorithm for computing the color grid for adversarial face mask attack is outlined in Algorithm~\ref{alg:face_mask}.
Here, we use the dodging attack on closed-set systems as an example. The algorithms for other settings are similar.
Note that $\bm{\delta}_T$ is the resulting color matrix, and the corresponding adversarial example is $\bm{x}+M\cdot\mathcal{T}(\bm{\delta}_T)$.  

\section{Universal Attack}

\subsection{Optimization Formulation}
The formulations of universal perturbations are presented in Table~\ref{tab:universal}.
In \model, we mainly focus on universal dodging attacks.
Effective universal impersonation attack is still an open problem, and we leave it for future work.

\subsection{Computing Universal Perturbations}

\begin{algorithm}[t]     
\begin{algorithmic}[1]  
\REQUIRE Target system $S$;\\ Input face image batch $\{\bm{x}_i, y_i\}_{i=1}^N$; \\
The number of iterations $T$;\\ Step size $\alpha$;\\ Momentum parameter $\mu$. 
\ENSURE The universal perturbation $\bm{\delta}_T$ for $\{\bm{x}_i, y_i\}_{i=1}^N$.
\STATE Initialize $\bm{\delta}_0 := \bm{0}$, $\bm{g}_0 := \bm{0}$;
\FOR{each $t \in [0, T-1]$}
\FOR{each $i \in [1, N]$}
\STATE $\ell_{i,t} := \ell(S(\bm{x}_i+M\cdot\bm{\delta}_t), y_i)$;
\ENDFOR
\STATE $\ell_t = \min\{\ell_{i,t}\}_{i=1}^N$;
\STATE $\bm{g}_{t+1}:=\mu \cdot \bm{g}_t + \frac{\nabla_{\bm{\delta}_t}\ell_t}{||\ell_t||_1}$;
\STATE $\bm{\delta}_{t+1}:= \bm{\delta}_t + \alpha\cdot\mathrm{sign}(\bm{g}_{t+1})$;
\STATE Clip $\bm{\delta}_{t+1}$ such that pixel values of $\bm{x}+M\cdot\bm{\delta}_{t+1}$ are in $[0, 255/255]$;
\ENDFOR  
\RETURN $\bm{\delta}_T$.  
\end{algorithmic}
\caption{Finding universal perturbations.}
\label{alg:universal} 
\end{algorithm}

\begin{figure*}[t]
\centering
\begin{tabular}{cc}
  \includegraphics[width=0.42\textwidth]{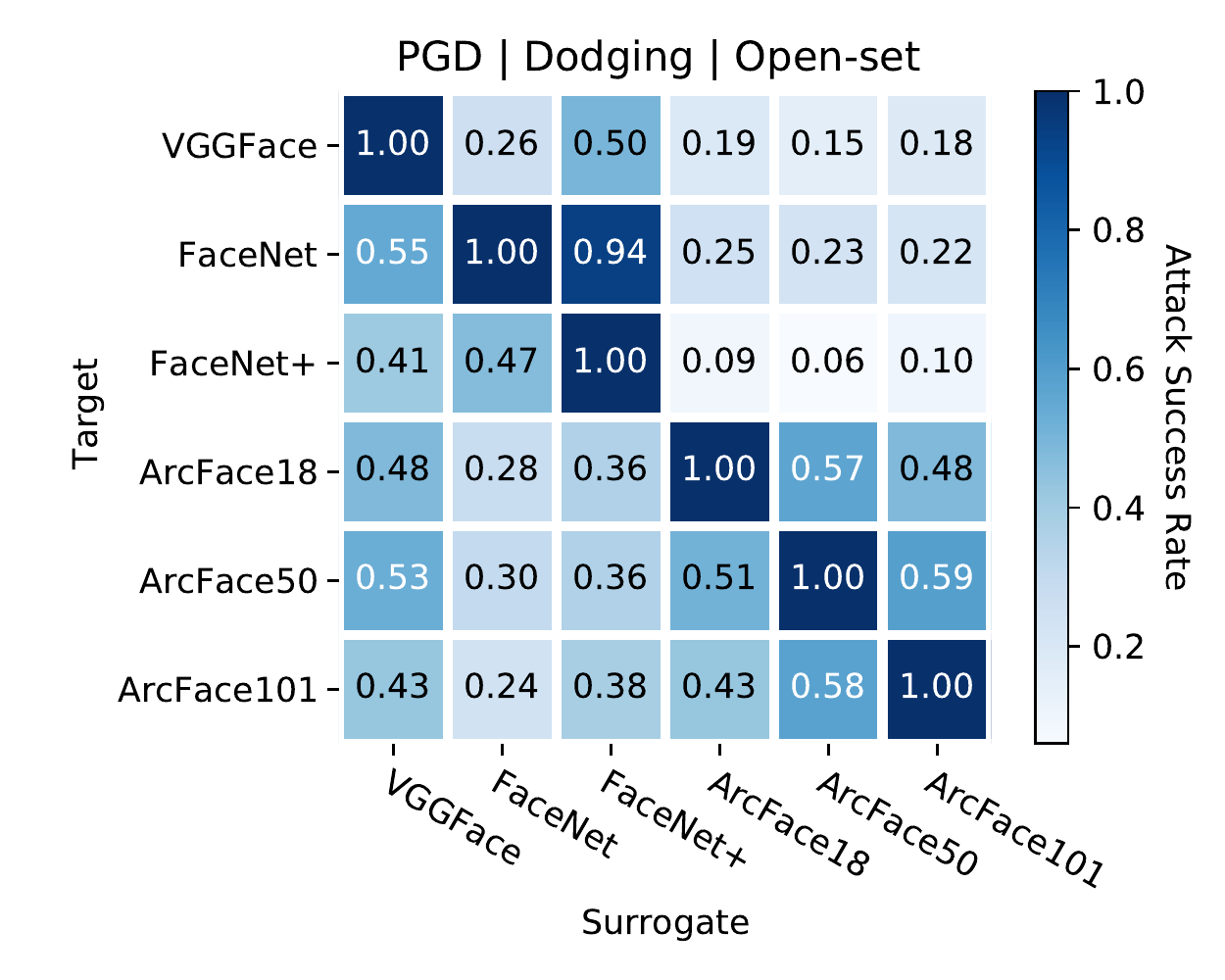} &
  \includegraphics[width=0.42\textwidth]{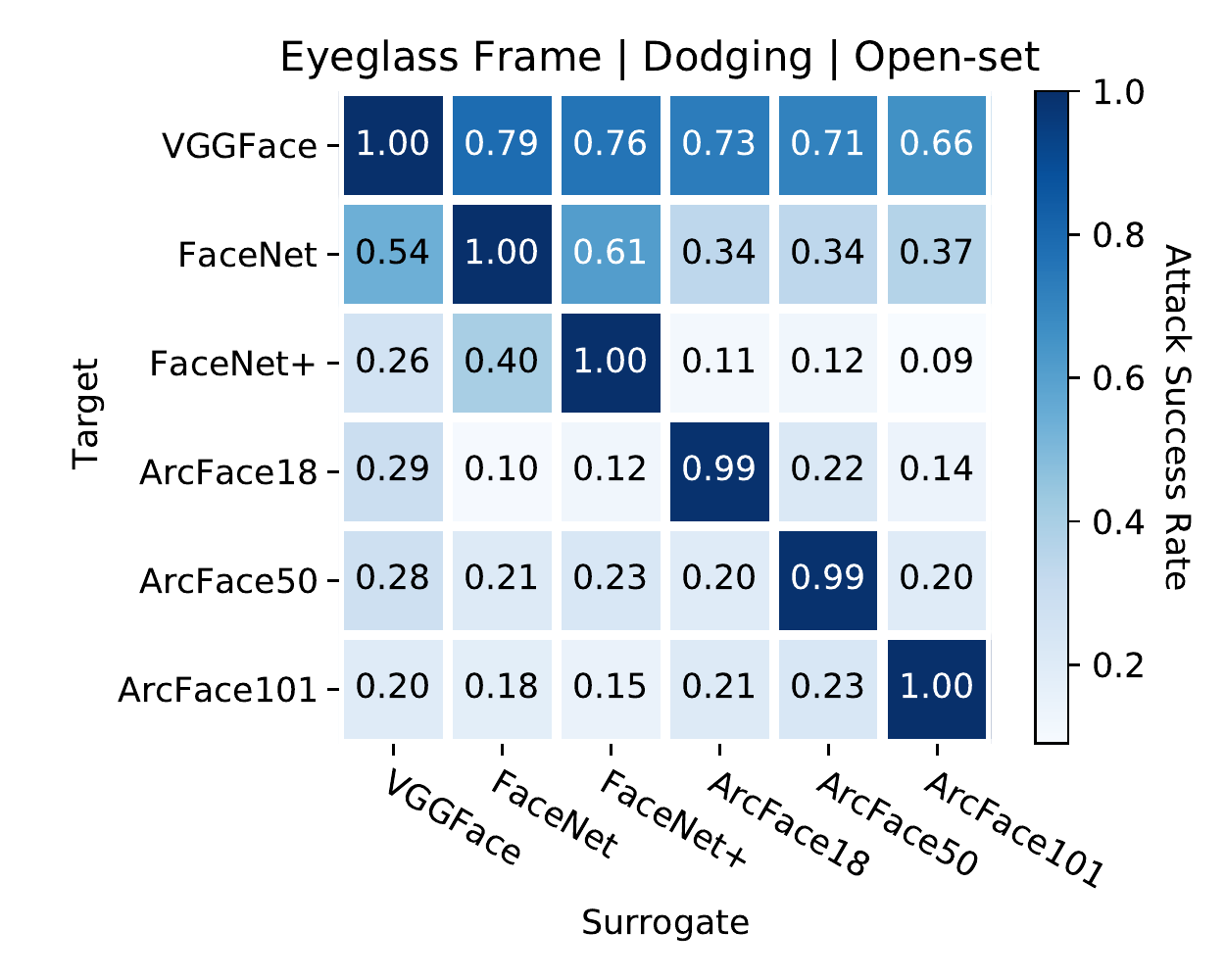}\\
  \includegraphics[width=0.42\textwidth]{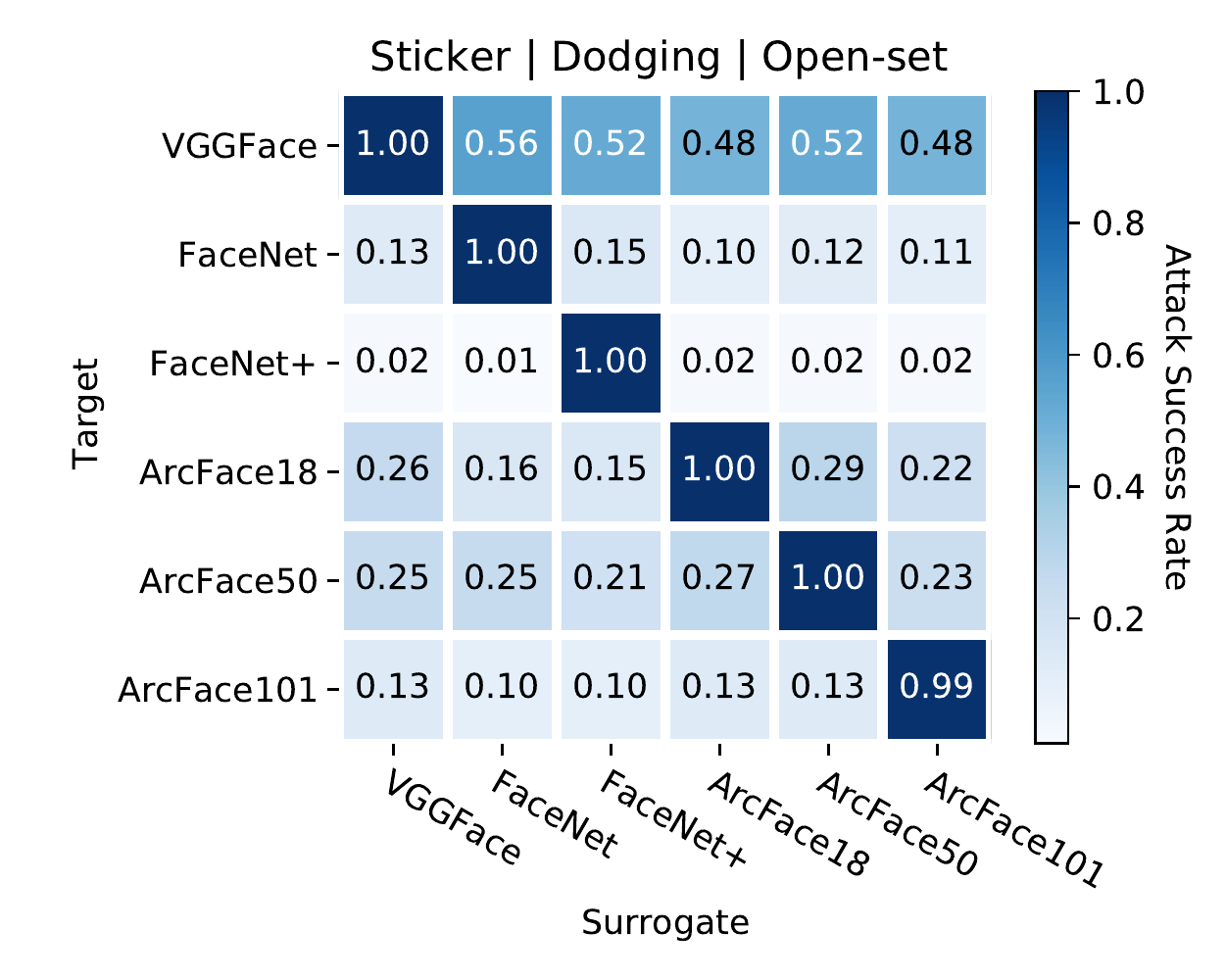} &
  \includegraphics[width=0.42\textwidth]{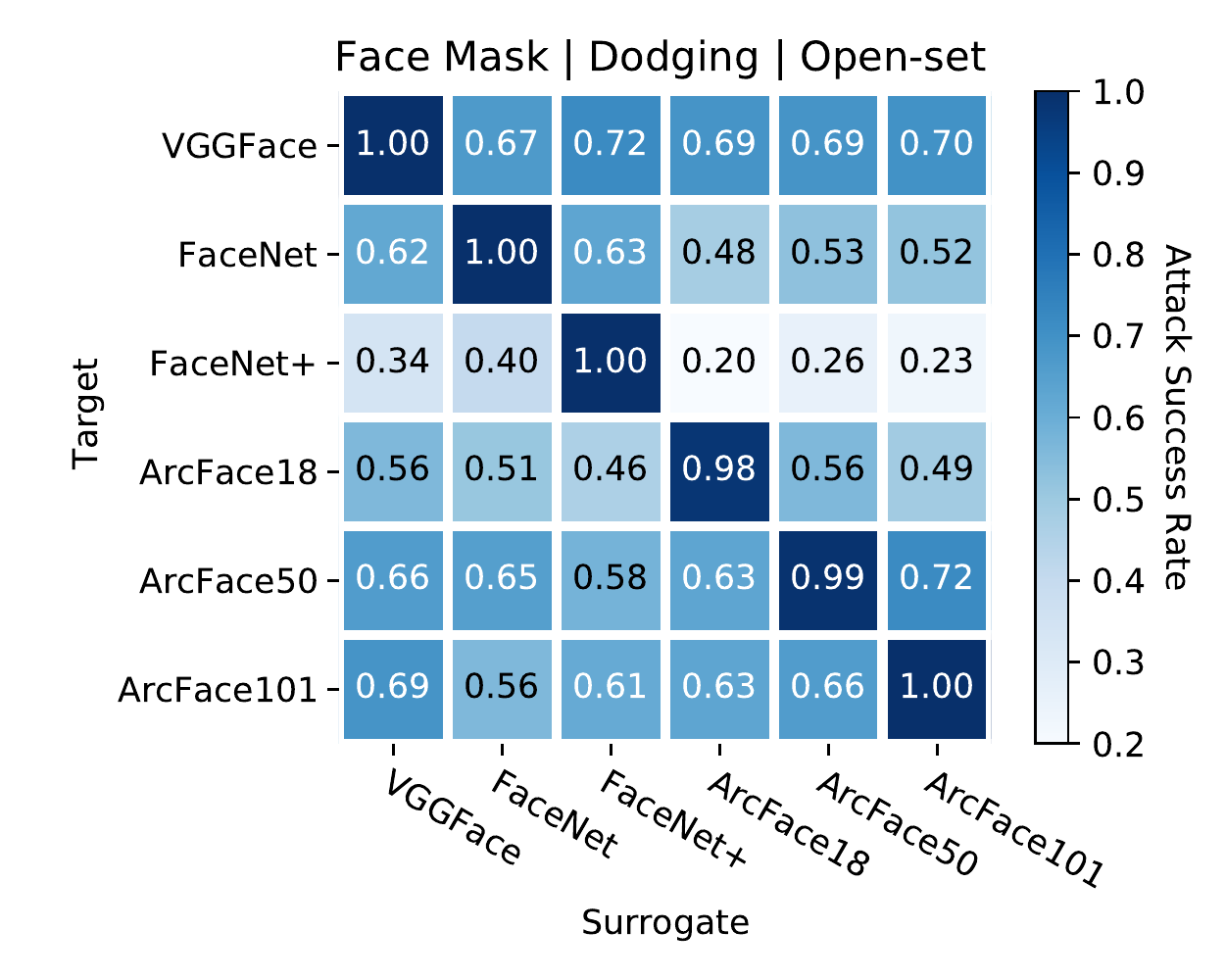} \\
\end{tabular}
\caption{Attack success rate of dodging attacks with different open-set targets and surrogate models.
Upper left: PGD attack.
Upper right: Eyeglass frame attack.
Lower left: Sticker attack.
Lower right: Face mask attack.
}
\label{fig:dodging_components_open}
\end{figure*}

The algorithm for finding universal perturbations is presented in Algorithm~\ref{alg:universal}.
Here, we use the dodging attack on closed-set systems as an example.
The algorithms for other settings are similar.
Note that in practice, 
Line 3--6 in Algorithm~\ref{alg:universal} can be executed in a paralleled manner by using GPUs.
Therefore, compared to traditional methods that iterate every data point to find a universal perturbation~\cite{moosavi2017universal}, our approach can achieve a significant speedup. 

\section{Robustness of Face Recognition Components}

\subsection{Open-set Systems Under Dodging Attacks}

To study the robustness of open-set system components under dodging attacks, we employ six different face recognition systems and then evaluate the attack success rates of dodging attacks corresponding to different target and surrogate face recognition models. 
Specifically, besides the five systems (VGGFace, FaceNet, ArcFace18, ArcFace50, and ArcFace101) presented in Table 2 of the main paper, we build a face recognition model by training FaceNet~\cite{schroff2015facenet} using the VGGFace2 dataset~\cite{cao2018vggface2} (henceforth, \emph{FaceNet+}).
Here, FaceNet and FaceNet+ are trained using the same neural architecture but different training sets, while the ArcFace variations share the same training data but with different architectures.
The results are presented in Fig.~\ref{fig:dodging_components_open}.

We have the following two observations, which are similar to those observed from dodging attacks on closed-set systems in the main paper.
First, in most cases, an open-set system's neural architecture is more fragile than its training set.
For example, under the PGD attack, adversarial examples in response to FaceNet+ have a $94\%$ success rate on FaceNet (which is trained using the same architecture but different training data), while the success rates among the ArcFace systems (which are built with the same training set but different neural architectures) are only around $50\%$. However, there are also some cases where the neural architecture exhibits similar robustness to the training set. 
For example, when black-box attacks are too weak (under sticker attack), both neural architecture and training set are robust; when the attacks are too strong (under face mask attack), these two components exhibit similar levels of vulnerability.
Second, the grid-level face mask attack is considerably more effective than the PGD attack, and significantly more potent than other physically realizable attacks.  
Like dodging attacks in closed-set settings, most black-box pixel-level physically realizable attacks have relatively low transferability on open-set face recognition systems, with only about $20\%$ success rate.

\begin{table}[]
\centering
\caption{Attack success rate of impersonation attacks on closed-set face recognition systems by the attacker's system knowledge. Z represents zero knowledge, T is training set, A is neural architecture, and F represents full knowledge.}
\scalebox{0.8}{
\begin{tabular}{|c|c|c|c|c|c|}
\hline
\multirow{2}{*}{\textbf{Target System}} & \multirow{2}{*}{\textbf{Attack Type}} & \multicolumn{4}{c|}{\textbf{Attacker's System Knowledge}} \\ \cline{3-6} 
                                        &                                       & \textbf{Z}   & \textbf{T}   & \textbf{A}   & \textbf{F}   \\ \hline
\multirow{4}{*}{VGGFace}                & PGD                                   & 0.11         & 0.21         & 0.35         & 1.00         \\ \cline{2-6} 
                                        & Eyeglass Frame                        & 0.01         & 0.01         & 0.03         & 0.95         \\ \cline{2-6} 
                                        & Sticker                               & 0.00         & 0.00         & 0.00         & 1.00         \\ \cline{2-6} 
                                        & Face Mask                             & 0.00         & 0.01         & 0.02         & 1.00         \\ \hline\hline
\multirow{4}{*}{FaceNet}                & PGD                                   & 0.23         & 0.32         & 1.00         & 1.00         \\ \cline{2-6} 
                                        & Eyeglass Frame                        & 0.00         & 0.00         & 0.28         & 0.99         \\ \cline{2-6} 
                                        & Sticker                               & 0.01         & 0.00         & 0.21         & 1.00         \\ \cline{2-6} 
                                        & Face Mask                             & 0.00         & 0.00         & 0.26         & 0.99         \\ \hline\hline
\multirow{4}{*}{ArcFace18}              & PGD                                   & 0.18         & 0.25         & 0.69         & 1.00         \\ \cline{2-6} 
                                        & Eyeglass Frame                        & 0.01         & 0.01         & 0.05         & 0.89         \\ \cline{2-6} 
                                        & Sticker                               & 0.00         & 0.00         & 0.01         & 0.94         \\ \cline{2-6} 
                                        & Face Mask                              & 0.01         & 0.01         & 0.03         & 0.77         \\ \hline\hline
\multirow{4}{*}{ArcFace50}              & PGD                                   & 0.13         & 0.15         & 0.45         & 0.87         \\ \cline{2-6} 
                                        & Eyeglass Frame                        & 0.02         & 0.02         & 0.03         & 0.67         \\ \cline{2-6} 
                                        & Sticker                               & 0.00         & 0.00         & 0.00         & 0.58         \\ \cline{2-6} 
                                        & Face Mask                             & 0.01         & 0.01         & 0.01         & 0.60         \\ \hline\hline
\multirow{4}{*}{ArcFace101}             & PGD                                   & 0.14         & 0.16         & 0.42         & 0.96         \\ \cline{2-6} 
                                        & Eyeglass Frame                        & 0.00         & 0.00         & 0.03         & 0.58         \\ \cline{2-6} 
                                        & Sticker                               & 0.00         & 0.00         & 0.00         & 0.50         \\ \cline{2-6} 
                                        & Face Mask                             & 0.01         & 0.01         & 0.04         & 0.73         \\ \hline
\end{tabular}
}
\label{tab:component-impersonation-closed}
\end{table}

\begin{figure*}[t]
\centering
\begin{tabular}{cc}
  \includegraphics[width=0.42\textwidth]{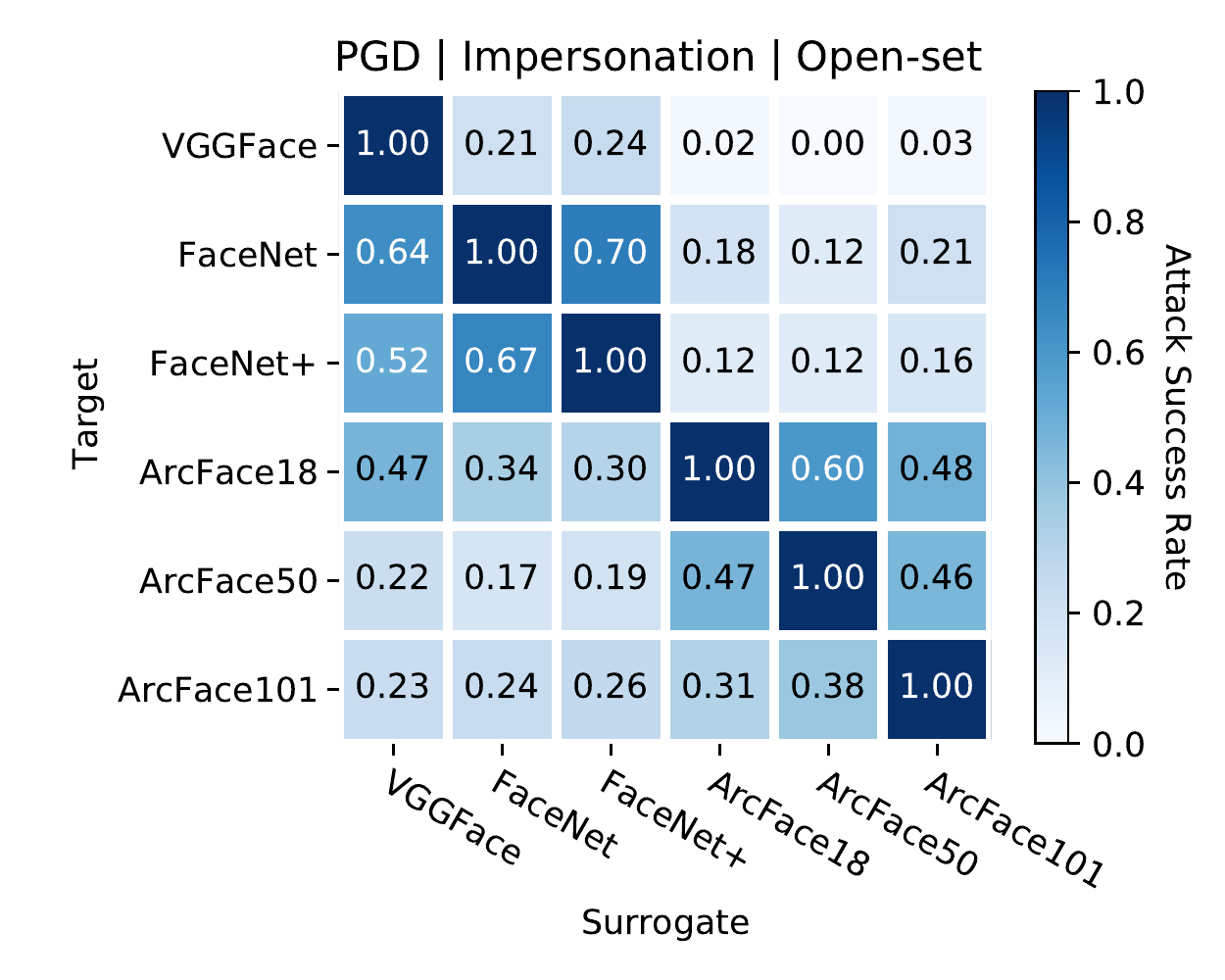} &
  \includegraphics[width=0.42\textwidth]{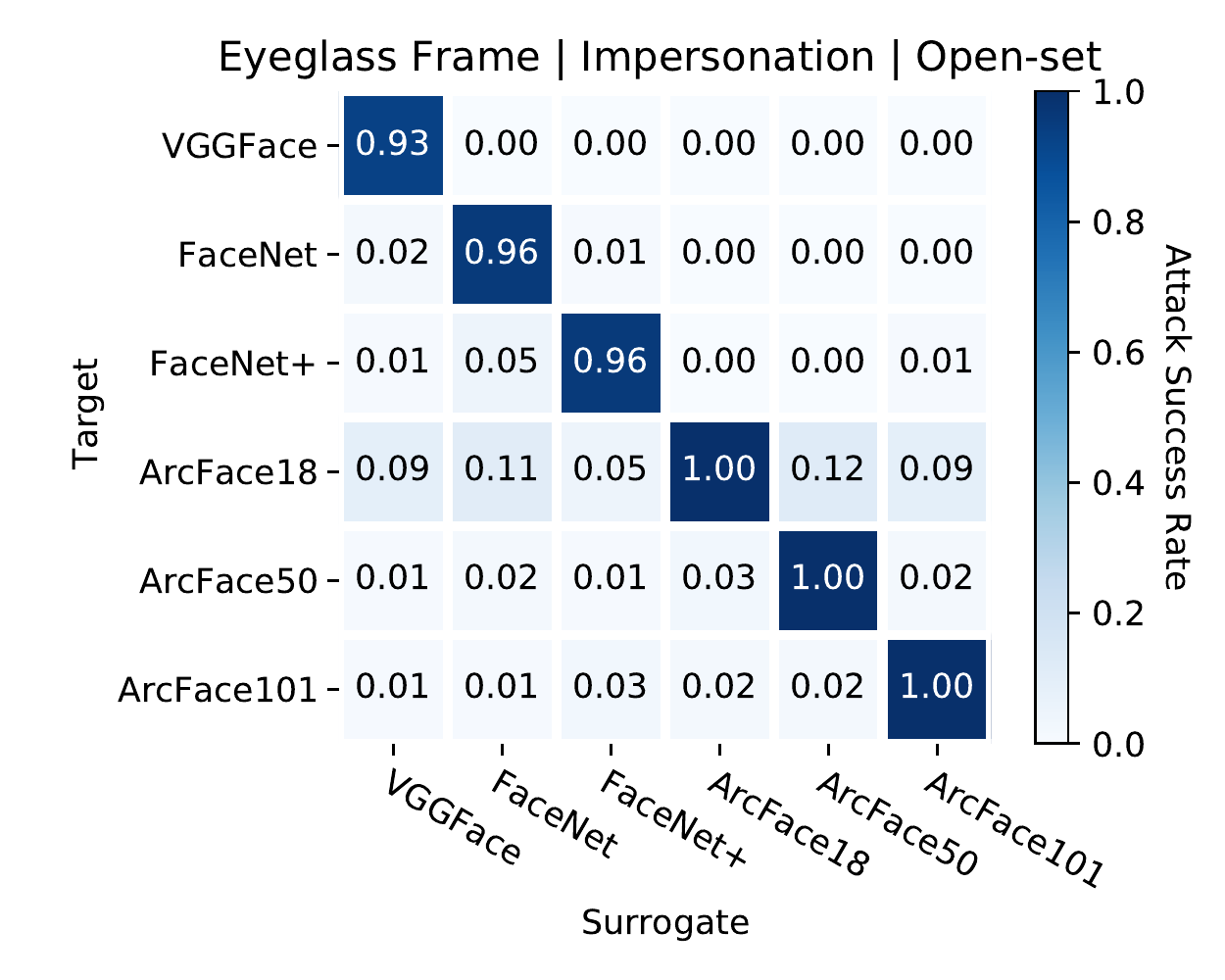}\\
  \includegraphics[width=0.42\textwidth]{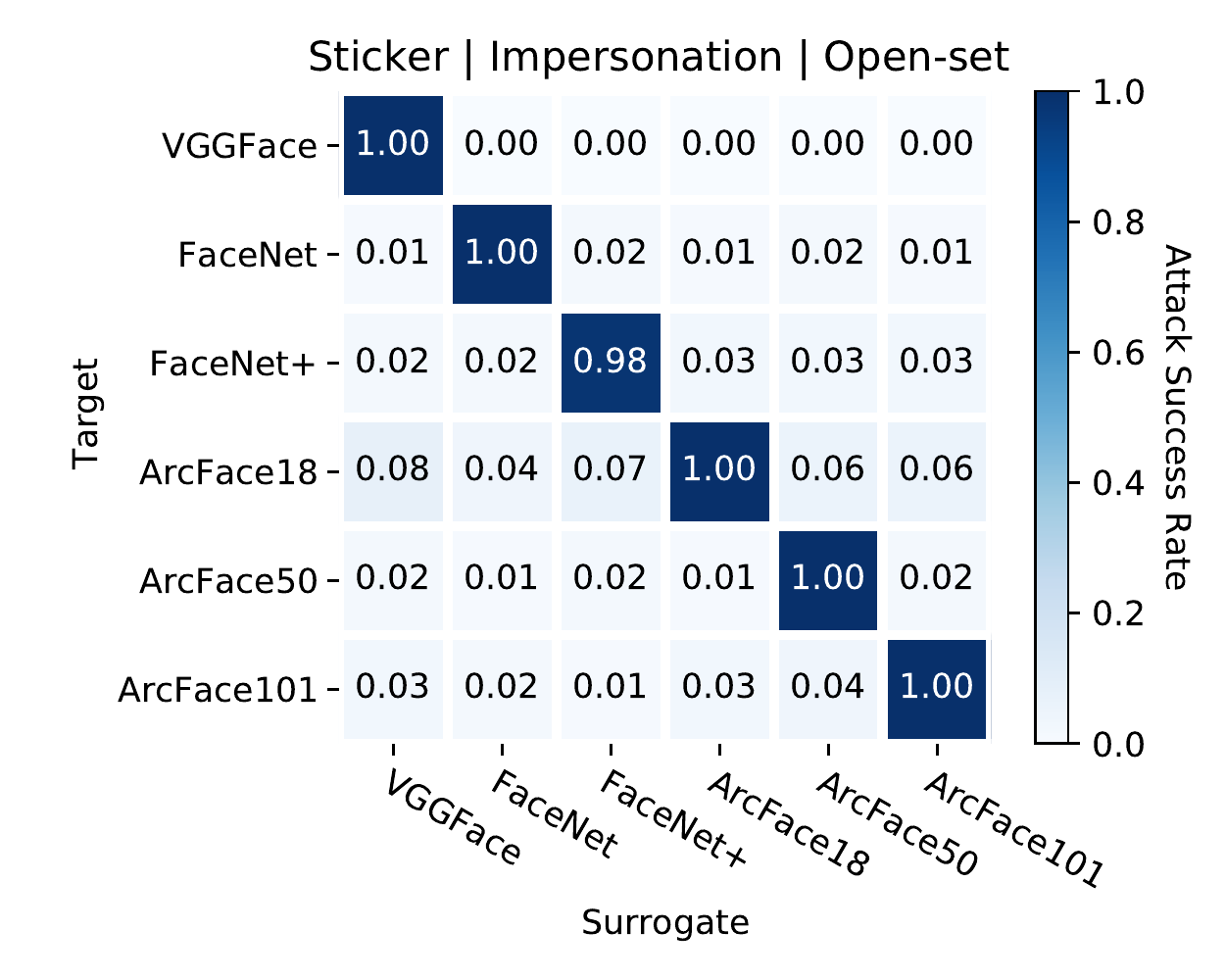} &
  \includegraphics[width=0.42\textwidth]{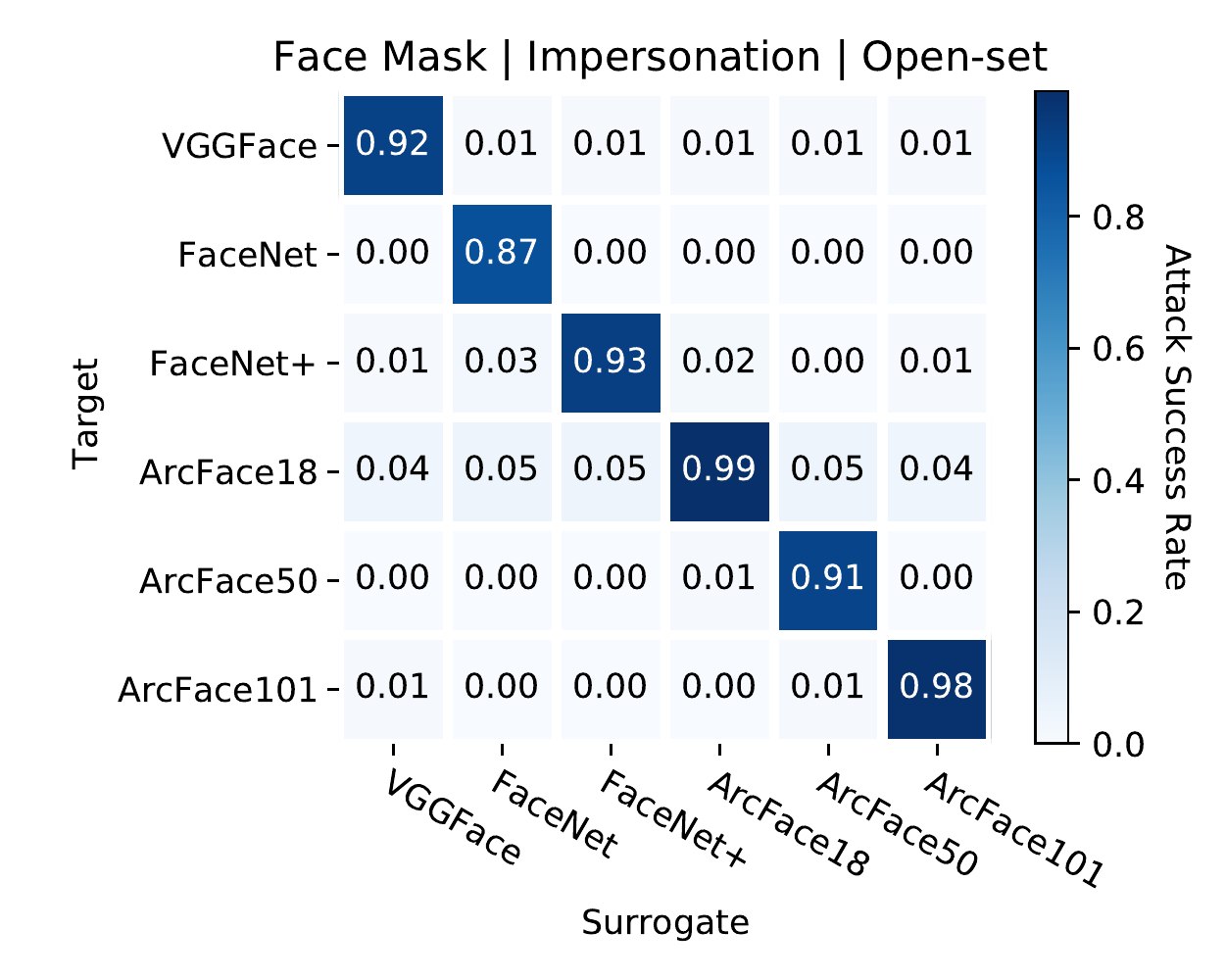} \\
\end{tabular}
\caption{Attack success rate of impersonation attacks with different open-set targets and surrogate models.
Upper left: PGD attack.
Upper right: Eyeglass frame attack.
Lower left: Sticker attack.
Lower right:Face mask attack.
}
\label{fig:impersonation_components_open}
\end{figure*}

\subsection{Closed-set Systems Under Impersonation Attacks}

Here, we use impersonation attacks to evaluate the robustness of closed-set systems.
In our experiments, all the closed-set models are 100-class classifiers, as introduced in Section 4.1 of the main paper.  
For any input face image $\bm{x}$ and its identity $y\in [0, 99]$, we let the target identity of the impersonation attack to be $y_t = (y+1)\%100$.
An impersonation attack is successful only when the resulting adversarial example is misclassified as the target identity $y_t$.
The results are shown in Table~\ref{tab:component-impersonation-closed}. 

We have two key findings.
First, compared to Table 3 of the main paper, we observe that closed-set systems are significantly more robust to impersonation attacks than dodging attacks.
Especially when an attacker has no accurate knowledge about the target system, the attack success rate of physically realizable attacks can be as low as $0\%$.
Second, it can be seen that closed-set systems exhibit moderate robustness against digital impersonation attacks. 
In such attacks, the knowledge of neural architecture is significantly more important than the training set.
For example, by knowing the neural architecture of ArcFace18, a PGD attack can achieve a $69\%$ success rate. In contrast, this rate drops to $25\%$ when only the training set is visible to the attacker. 

\subsection{Open-set Systems Under Impersonation Attacks}

To evaluate impersonation attacks on open-set systems, we randomly select 100 pairs from the LFW dataset~\cite{LFWTech} in a way similar to Section 4.1 of the main paper.
Each pair contains two face images corresponding to different identities. 
We let one image as the input $\bm{x}$ and the other as the target gallery image $\bm{x}^*_t$.
An impersonation attack is successful only when the resulting adversarial example and $\bm{x}^*_t$ are verified as the same identity.
The experimental results are presented in Fig.~\ref{fig:impersonation_components_open}.

Similar to the impersonation attacks on closed-set systems, we have the following observations that are consistent with our previous summary.
First, open-set systems are very robust to black-box impersonation physically realizable attacks.
In most cases, these attacks can only achieve a success rate of less than $10\%$.
In contrast, the PGD attack is significantly more potent. And under this attack, the neural architecture is considerably more vulnerable than the training set (\textit{e.g.}, comparing FaceNet variations to ArcFace models).

\section{Efficacy of Momentum and Ensemble Models in Transfer-based Attacks}

Next, we evaluate the efficacy of using momentum and ensemble-based surrogate models in transfer-based dodging attacks.
For a given closed-set target face recognition system, we first train a surrogate model using the same training data.
Specifically, we use both a \emph{single} surrogate trained on a different architecture\footnote{For a given target model, we trained four single surrogates corresponding to the other four architectures. Below, we only present the result of the surrogate that has the highest attack success rate.}, and an \emph{ensembled} surrogate by ensembling the other four systems in the way described in Section 3.2 of the main paper. 
We then produce white-box dodging attacks on the surrogate and evaluate the resulting examples' attack success rate on the target model.
For each attack, we compare the momentum method (\textit{i.e.}, w/ mmt) and the conventional gradient-based approach (\textit{i.e.}, w/o mmt).
The results are shown in Table~\ref{tab:transfer-pgd},~\ref{tab:transfer-eyeglass},~\ref{tab:transfer-sticker}, and~\ref{tab:transfer-face}.

We have two key observations.
First, both ensemble and momentum contribute to stronger transferability, although in most cases, ensemble contributes more.
For example, the ensemble method can boost the transferability of PGD attacks on FaceNet by $31\%$, while the improvement by momentum is only about $10\%$.
Second, the efficacy of momentum and ensemble models is highly dependent on the nature of perturbation.
For digital attacks, these methods combined can significantly improve transferability by up to $55\%$.  
In grid-level face mask attacks, the improvement is as considerable as up to $16\%$.
However, both methods can only marginally boost the transferability of pixel-level realizable attacks.
Especially in the sticker attacks, the improvement is nearly negligible. 
We leave effective transfer-based pixel-level physically realizable attacks as an open problem for future research.

\begin{table}[t]
\centering
\caption{Attack success rate of dodging PGD attacks on closed-set face recognition systems.
Here, only the target system's training data is visible to the attacker, and we use different surrogate models.}
\scalebox{0.85}{
\begin{tabular}{|c|c|c|c|c|}
\hline
\multirow{3}{*}{\textbf{Target System}} & \multicolumn{4}{c|}{\textbf{Surrogate System}}                                \\ \cline{2-5} 
                                        & \multicolumn{2}{c|}{\textbf{Single}} & \multicolumn{2}{c|}{\textbf{Ensemble}} \\ \cline{2-5} 
                                        & \textbf{w/o mmt}  & \textbf{w/ mmt}  & \textbf{w/o mmt}   & \textbf{w/ mmt}   \\ \hline\hline
VGGFace                                 & 0.08              & 0.16             & 0.43               & 0.51              \\ \hline
FaceNet                                 & 0.42              & 0.52             & 0.73               & 0.83              \\ \hline
ArcFace18                               & 0.42              & 0.51             & 0.87               & 0.92              \\ \hline
ArcFace50                               & 0.35              & 0.55             & 0.86               & 0.90              \\ \hline
ArcFace101                              & 0.32              & 0.39             & 0.71               & 0.78              \\ \hline
\end{tabular}
}
\label{tab:transfer-pgd}
\end{table}

\begin{table}[t]
\centering
\caption{Attack success rate of dodging eyeglass frame attacks on closed-set face recognition systems.
Here, only the target system's training data is visible to the attacker, and we use different surrogate models.}
\scalebox{0.85}{
\begin{tabular}{|c|c|c|c|c|}
\hline
\multirow{3}{*}{\textbf{Target System}} & \multicolumn{4}{c|}{\textbf{Surrogate System}}                                \\ \cline{2-5} 
                                        & \multicolumn{2}{c|}{\textbf{Single}} & \multicolumn{2}{c|}{\textbf{Ensemble}} \\ \cline{2-5} 
                                        & \textbf{w/o mmt}  & \textbf{w/ mmt}  & \textbf{w/o mmt}   & \textbf{w/ mmt}   \\ \hline\hline
VGGFace                                 & 0.17              & 0.22             & 0.26               & 0.28              \\ \hline
FaceNet                                 & 0.08              & 0.09             & 0.14               & 0.16              \\ \hline
ArcFace18                               & 0.02              & 0.03             & 0.05               & 0.06              \\ \hline
ArcFace50                               & 0.05              & 0.05             & 0.10               & 0.12              \\ \hline
ArcFace101                              & 0.02              & 0.03             & 0.02               & 0.03              \\ \hline
\end{tabular}
}
\label{tab:transfer-eyeglass}
\end{table}

\begin{table}[t]
\centering
\caption{Attack success rate of dodging sticker attacks on closed-set face recognition systems.
Here, only the target system's training data is visible to the attacker, and we use different surrogate models.}
\scalebox{0.85}{
\begin{tabular}{|c|c|c|c|c|}
\hline
\multirow{3}{*}{\textbf{Target System}} & \multicolumn{4}{c|}{\textbf{Surrogate System}}                                \\ \cline{2-5} 
                                        & \multicolumn{2}{c|}{\textbf{Single}} & \multicolumn{2}{c|}{\textbf{Ensemble}} \\ \cline{2-5} 
                                        & \textbf{w/o mmt}  & \textbf{w/ mmt}  & \textbf{w/o mmt}   & \textbf{w/ mmt}   \\ \hline\hline
VGGFace                                 & 0.02              & 0.02             & 0.06               & 0.06              \\ \hline
FaceNet                                 & 0.00              & 0.00             & 0.01               & 0.01              \\ \hline
ArcFace18                               & 0.00              & 0.00             & 0.01               & 0.01              \\ \hline
ArcFace50                               & 0.00              & 0.00             & 0.00               & 0.01              \\ \hline
ArcFace101                              & 0.00              & 0.01             & 0.04               & 0.04              \\ \hline
\end{tabular}
}
\label{tab:transfer-sticker}
\end{table}

\begin{table}[t]
\centering
\caption{Attack success rate of dodging face mask attacks on closed-set face recognition systems.
Here, only the target system's training data is visible to the attacker, and we use different surrogate models.}
\scalebox{0.85}{
\begin{tabular}{|c|c|c|c|c|}
\hline
\multirow{3}{*}{\textbf{Target System}} & \multicolumn{4}{c|}{\textbf{Surrogate System}}                                \\ \cline{2-5} 
                                        & \multicolumn{2}{c|}{\textbf{Single}} & \multicolumn{2}{c|}{\textbf{Ensemble}} \\ \cline{2-5} 
                                        & \textbf{w/o mmt}  & \textbf{w/ mmt}  & \textbf{w/o mmt}   & \textbf{w/ mmt}   \\ \hline\hline
VGGFace                                 & 0.18              & 0.26             & 0.20               & 0.32              \\ \hline
FaceNet                                 & 0.26              & 0.38             & 0.42               & 0.42              \\ \hline
ArcFace18                               & 0.21              & 0.33             & 0.21               & 0.33              \\ \hline
ArcFace50                               & 0.28              & 0.34             & 0.36               & 0.36              \\ \hline
ArcFace101                              & 0.22              & 0.34             & 0.30               & 0.36              \\ \hline
\end{tabular}
}
\label{tab:transfer-face}
\end{table}

\begin{table}[t]
\centering
\caption{Attack success rate of dodging attacks on open-set face recognition systems by the universality of adversarial examples.
Here, $N$ represents the batch size of face images that share a universal perturbation. 
}
\scalebox{0.8}{
\begin{tabular}{|c|c|c|c|c|c|}
\hline
\multirow{2}{*}{\textbf{Target System}} & \multirow{2}{*}{\textbf{Attack Type}} & \multicolumn{4}{c|}{\textbf{Attacker's Capability}} \\ \cline{3-6} 
                                        &                                       & \textbf{N=1}   & \textbf{N=5}   & \textbf{N=10}   & \textbf{N=20}   \\ \hline
\multirow{4}{*}{VGGFace}                & PGD                                   & 1.00   & 0.89   & 0.81   & 0.53         \\ \cline{2-6} 
                                        & Eyeglass Frame                        & 1.00   & 1.00   & 1.00   & 1.00         \\ \cline{2-6} 
                                        & Sticker                               & 1.00   & 1.00   & 1.00   & 1.00         \\ \cline{2-6} 
                                        & Face Mask                             & 1.00   & 1.00   & 1.00   & 1.00         \\ \hline\hline
\multirow{4}{*}{FaceNet}                & PGD                                   & 1.00   & 0.02   & 0.02   & 0.02         \\ \cline{2-6} 
                                        & Eyeglass Frame                        & 1.00   & 1.00   & 1.00   & 1.00         \\ \cline{2-6} 
                                        & Sticker                               & 1.00   & 1.00   & 0.99   & 0.90         \\ \cline{2-6} 
                                        & Face Mask                             & 1.00   & 1.00   & 0.99   & 0.98          \\ \hline\hline
\multirow{4}{*}{ArcFace18}              & PGD                                   & 1.00   & 0.96   & 0.79   & 0.46         \\ \cline{2-6} 
                                        & Eyeglass Frame                        & 0.99   & 0.86   & 0.70   & 0.67         \\ \cline{2-6} 
                                        & Sticker                               & 1.00   & 1.00   & 1.00   & 0.99         \\ \cline{2-6} 
                                        & Face Mask                             & 0.98   & 0.98   & 0.93   & 0.92         \\ \hline\hline
\multirow{4}{*}{ArcFace50}              & PGD                                   & 1.00   & 0.91   & 0.75   & 0.47         \\ \cline{2-6} 
                                        & Eyeglass Frame                        & 0.99   & 0.78   & 0.67   & 0.62         \\ \cline{2-6} 
                                        & Sticker                               & 1.00   & 1.00   & 1.00   & 0.00         \\ \cline{2-6} 
                                        & Face Mask                             & 0.99   & 0.99   & 0.99   & 0.94         \\ \hline\hline
\multirow{4}{*}{ArcFace101}             & PGD                                  & 1.00   & 0.68   & 0.68   & 0.41         \\ \cline{2-6} 
                                        & Eyeglass Frame                       & 1.00   & 0.85   & 0.73   & 0.65         \\ \cline{2-6} 
                                        & Sticker                              & 0.99   & 0.98   & 0.97   & 0.97         \\ \cline{2-6} 
                                        & Face Mask                            & 1.00   & 1.00   & 1.00   & 1.00         \\ \hline
\end{tabular}
}
\label{tab:universal-open}
\end{table}

\section{Universal Attacks}

Finally, we evaluate open-set systems under universal dodging attacks.
The results are shown in Table~\ref{tab:universal-open}.
Compared to Table 5 of the main paper, we find that open-set systems are significantly more fragile to universal perturbations of all types than their closed-set counterparts.
For example, when $N=20$, the open-set ArcFace101 is susceptible to all the four types of universal attacks, while in the closed-set setting it is only vulnerable to the universal face mask attack. 
Moreover, we again observe that the universal grid-level face mask attack is more effective than the other perturbation types.
Here, we also find that the sticker attack is as potent as the face mask attack in open-set settings.

\end{document}